\documentclass{article}

\usepackage[eandd, final]{neurips_2026}

\usepackage[utf8]{inputenc} % allow utf-8 input
\usepackage[T1]{fontenc}    % use 8-bit T1 fonts
\usepackage{hyperref}       % hyperlinks
\hypersetup{
  pdftitle={Realism VS Accuracy: Event Sequence Forecasting from a Generative Modeling Perspective},
  pdfauthor={Dmitry Osin, Egor Surkov, Petr Mokrov, Igor Udovichenko, Ivan Karpukhin, Evgeny Burnaev, Egor Shvetsov, Alexander Korotin}
}
\usepackage{url}            % simple URL typesetting
\usepackage{booktabs}       % professional-quality tables
\usepackage{amsfonts}       % blackboard math symbols
\usepackage{nicefrac}       % compact symbols for 1/2, etc.
\usepackage{microtype}      % microtypography
\usepackage[table]{xcolor}         % colors
\usepackage{amsmath, amssymb, amsthm}
\usepackage{graphicx}
\usepackage{subcaption}
\usepackage{bbding}
\usepackage{pifont}
\usepackage{enumitem}
\usepackage{adjustbox}
\usepackage{makecell}
\usepackage{array}          % p{}, m{}, b{} columns and >{} hooks
\usepackage{float}          % H placement specifier
\usepackage{algorithm}
\usepackage{algpseudocode}
\usepackage[normalem]{ulem} % \uline, \dashuline, \dotuline (normalem keeps \emph intact)
\usepackage[most]{tcolorbox}
\usepackage{etoc}

\newcommand{\defeq}{\stackrel{\text{def}}{=}}

\newcommand{\bk}{\mathbf{k}}

\newcommand{\bx}{\mathbf{x}}

\newcommand{\bbE}{\mathbb{E}}

\newcommand{\bbR}{\mathbb{R}}

\newcommand{\cmark}{{\setlength{\fboxsep}{1pt}\colorbox{green!60!black}{\textcolor{white}{\ding{51}}}}}
\newcommand{\xmark}{\textcolor{red}{\ding{56}}}
\newcommand{\med}[1]{\mathrm{med}\left[#1\right]}

\newif\ifcleancopy % false by default: revisions are shown with styling

\newcommand{\cleancopy}{\cleancopytrue}

\newtcolorbox{semitransparentsectionbox}{
  enhanced,
  breakable,
  boxrule=0.4pt,
  arc=2pt,
  colframe=cyan!35!black,
  colback=cyan!10!white,
  opacityback=0.9,
  left=6pt,
  right=6pt,
  top=6pt,
  bottom=6pt
}
\title{Realism VS Accuracy: \\Event Sequence Forecasting \\from a Generative Modeling Perspective}

\author{%
  \textbf{Dmitry Osin}\textsuperscript{\mdseries 1,2}\thanks{Corresponding author: \texttt{dima.tina2013@gmail.com}} \quad
  \textbf{Egor Surkov}\textsuperscript{\mdseries 1} \quad
  \textbf{Petr Mokrov}\textsuperscript{\mdseries 1} \quad
  \textbf{Igor Udovichenko}\textsuperscript{\mdseries 1,3} \\
  \textbf{Ivan Karpukhin}\textsuperscript{4} \quad
  \textbf{Evgeny Burnaev}\textsuperscript{1,5} \quad
  \textbf{Egor Shvetsov}\textsuperscript{1} \quad
  \textbf{Alexander Korotin}\textsuperscript{1,5} \\[6pt]
  \textsuperscript{1}Applied AI Institute \quad
  \textsuperscript{2}ITMO University \quad
  \textsuperscript{3}Vega Institute Foundation \quad
  \textsuperscript{4}SB AI Lab \quad
  \textsuperscript{5}AXXX
}

\begin{document}
% The clean-copy build (removed text hidden, added text unstyled) is produced
% by compiling the driver file main-clean.tex, which defines \cleanflag and
% then \input{main}. See main-clean.tex.
\ifdefined\cleanflag\cleancopy\fi

\maketitle
\etocdepthtag.toc{main}

\begin{abstract}
Event sequences are chains of random events, each described by heterogeneous (numerical and categorical) features. 
The problem of event sequence forecasting occurs in finance, healthcare, business process intelligence, and other fields. 
In this work, we study multi-horizon forecasting of event sequences as a conditional generative modeling problem. 
We evaluate forecasts not only for prediction accuracy, but also for \textbf{distributional realism}.
To this end, we present a \textbf{benchmark} comprising metrics for both quality aspects, strong statistical baselines, recent models adapted from adjacent fields, and curated real-world datasets.
Careful evaluation reveals that no method excels at both accuracy and realism, highlighting a fundamental trade-off between the two.
Finally, we introduce \textbf{Wasserstein BaryBooster}, an algorithm that aggregates samples from a realism-strong diffusion model into a state-of-the-art predictor, demonstrating that distributional realism can be converted into prediction accuracy.
Code is available at \url{https://github.com/xaosina/event-sequence-forecasting}.
\end{abstract}

%-----------------------------------------------------------
\section{Introduction}
%-----------------------------------------------------------

\begin{figure}[t]
    \centering
    \includegraphics[width=\linewidth]{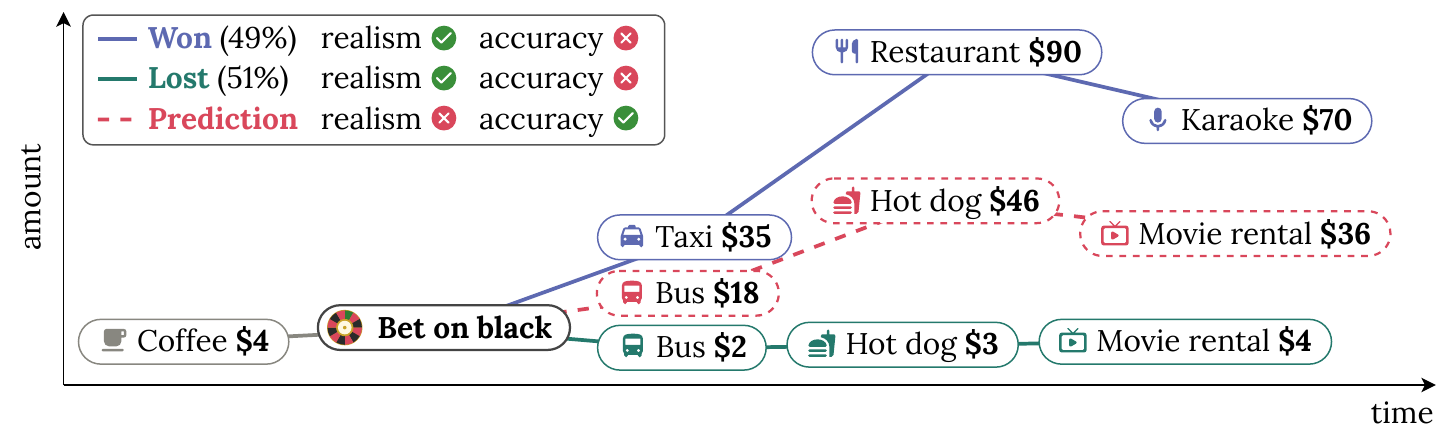}
    \caption{%
    A toy event sequence. After a bet on black at roulette, the evening follows one of two realistic chains of transactions: the person either lost (51\%) or won (49\%). The prediction that minimizes MSE and maximizes per-event accuracy takes the categories of the more likely chain and averages the amounts, pairing cheap categories with large sums (e.g., an \$18 bus ride). It scores best on predictive metrics, yet matches neither chain. On the other hand, each possible outcome is realistic, but its expected MSE is about twice as large.
    }
    \label{fig:intro-example}
\end{figure}

Many high-stakes decisions rest on modeling chains of future events: the next purchases on a customer's card~\citep{bazarova2025learning}, the upcoming entries in a patient's clinical record~\citep{mimic4}, the next actions of a player in an online game~\citep{zhao2023percltv}, and similar processes across other domains~\citep{lopes2019survey, liguori2025modeling}.
Each event is timestamped and described by a heterogeneous set of features --- typically a mix of numerical attributes (e.g., transaction amount) and categorical attributes (e.g., merchant category code).
We refer to such data as \emph{event sequences} (Figure~\ref{fig:intro-example}) and study the problem of forecasting them over long horizons given a recorded history.

The closest body of prior work is \emph{temporal point processes} (TPP), which model event timing~\citep{addthin, psd, editpp} and, in marked variants, a single discrete event type~\citep{easytpp, hotpp}.
While effective in their own setting, TPP methods do not natively handle the heterogeneous numerical and categorical features that real event sequences carry (see \S~\ref{sec:rw-models}).
To the best of our knowledge, prior work has not studied long-horizon forecasting of general event sequences systematically.

We frame event sequence forecasting as a \emph{conditional generative modeling} problem: given a history, the model produces samples that should be plausible both relative to that history and to the broader data distribution.
This framing implies two distinct perspectives on forecast quality.
\emph{Prediction accuracy} measures how close a forecast is to the actual continuation of the sequence.
\emph{Distributional realism} measures how indistinguishable forecasts are from genuine future trajectories drawn from the same distribution.
Prior evaluations, in TPP and beyond, have focused almost exclusively on prediction accuracy, overlooking realism.

To the best of our knowledge, no prior benchmark treats both dimensions as first-class citizens.
We make the following \textbf{contributions}:
\begin{enumerate}[leftmargin=*]
    \item We implement a suite of statistical baselines (\S\ref{sec:methods:baselines}) and adapt cutting-edge generative diffusion models and SoTA predictive models (\S\ref{sec:ldm}--\S\ref{sec:def}) to the task of forecasting event sequences with heterogeneous features.
    
    \item In \S\ref{sec:benchmark} we present a \textbf{benchmark} comprising several large-scale event sequence datasets and a set of metrics that capture both perspectives on forecast quality, including but not limited to \textbf{cDFS} (Conditional Detection Fooling Score), an adaptation of the classifier-two-sample test~\citep{c2st}, and \textbf{TMS} (Temporal Matching Score), our novel matching-based metric inspired by the optimal transport distance used extensively in TPP research~\citep{mei2019imputing, easytpp, hotpp}.
        
    \item We find (\S\ref{sec:results}) that no method excels at both TMS and cDFS, exposing a \textbf{fundamental trade-off between prediction accuracy and distributional realism}.
    
    \item In \S\ref{sec:barybooster} we introduce \textbf{Wasserstein BaryBooster}, an aggregation procedure that, when applied on top of samples from a diffusion model, produces SoTA prediction accuracy.
\end{enumerate}

%-----------------------------------------------------------
\section{Problem Statement and Related Work}\label{sec:related}
%-----------------------------------------------------------

We formalize the problem of event sequence forecasting as follows.
Let $t$ be the timestamp of an event, and let $\bx$ and $\bk$ denote its numerical and categorical feature vectors, respectively.
An event sequence is a sequence of events: $S = \left( E_1, \dots, E_n \right)$, where each event is a 3-tuple $E_i = (t_i, \bx_i, \bk_i)$.
We denote by $S_{m:l} = \left( E_m, \dots, E_l \right)$ a slice of the event sequence.
Given a history $S_{1:H}$, the goal is to forecast the \emph{sequence} of future events $S_{H+1:H+L}$.

We position our paper among three relevant bodies of work: generative models applicable to event-sequence data (\S\ref{sec:rw-models}), existing benchmarks in the space (\S\ref{sec:rw-benchmarks}), and metrics previously used to evaluate such models (\S\ref{sec:rw-metrics}).
 
\subsection{Generative Models for Event Data}
\label{sec:rw-models}
 
\begin{table}[h]
\centering
\caption{%
    Comparison of related generative models.
    ``Features'' indicates the supported feature type: ``none'' (time only), ``cat'' (a single event type, as in marked TPP, MTPP), ``num'' (numerical features only, as in spatio-temporal TPP, STPP), or ``mixed'' (both numerical and categorical). 
    ``Seq.'' indicates whether the method natively handles sequential (as opposed to single-row tabular) data.
    ``Forecasting'' indicates whether the method allows history-conditional generation.
    ``Diffusion'' indicates whether the method is a latent diffusion model, a data-space diffusion model, or non-diffusion (ND).
    Methods applicable to event sequence forecasting are highlighted.
}
\label{tab:methods}
\begin{tabular}{lrccc}
\toprule
Method & Features & Seq. & Forecasting & Diffusion \\
\midrule
\rowcolor{green!20} CDiff~\citep{cdiff} & mixed \cmark & \cmark & \cmark & data \\
TabSyn~\citep{tabsyn} & mixed \cmark & \xmark & \xmark & latent \\
TabDiff~\citep{tabdiff} & mixed \cmark & \xmark & \xmark & data \\
Add and Thin~\citep{addthin} & none \xmark & \cmark & \cmark & data \\
PSDiff~\citep{psd} & num \xmark & \cmark & \cmark & data \\
EdiTPP~\citep{editpp} & none \xmark & \cmark & \cmark & data \\
\rowcolor{green!20} DEF~\citep{def} & mixed \cmark & \cmark & \cmark & ND \\
UniTTab~\citep{unittab} & mixed \cmark & \cmark & \xmark & ND \\
TabDiT~\citep{tabdit} & mixed \cmark & \cmark & \xmark & latent \\
\rowcolor{green!20} ADiff4TPP~\citep{adiff4tpp} & mixed \cmark & \cmark & \cmark & latent \\
\bottomrule
\end{tabular}
\end{table}
 
Event sequences consist of events described by both numerical and categorical features, along with their timestamps.
This structure places them at the intersection of three lines of generative-modeling work, which we summarize in Table~\ref{tab:methods}.
We characterize methods by (i) their model type --- \emph{latent} diffusion trained in a learned representation space, \emph{data} diffusion designed directly in feature space, or non-diffusion (ND); (ii) whether they natively handle sequential data; and (iii) the feature types they can generate.
Appendix~\ref{app:rw-models} discusses \uline{each of these three classes in detail} --- tabular-data generative models, temporal point processes, and models for general heterogeneous event sequences --- together with the specific methods that we adapt as baselines and evaluate throughout this benchmark.

\begin{semitransparentsectionbox}
Although many of these methods \emph{can be applied} to forecasting of event sequences with heterogeneous features, to the best of our knowledge, \emph{no prior work has done so}.
We fill this gap by adapting representatives of all three classes to the general event-sequence forecasting setting: a data-space diffusion model, a latent diffusion model, and a non-diffusion method.
\end{semitransparentsectionbox}

\subsection{Benchmarks for Event Sequences}
\label{sec:rw-benchmarks}
 
Published benchmarks are scarce.
The three closest to our work are EasyTPP~\citep{easytpp}, HoTPP~\citep{hotpp}, and EBES~\citep{ebes}.
We \uline{review each of them in detail} in Appendix~\ref{app:rw-benchmarks}; here we summarize how our benchmark relates to all three.

Our benchmark occupies a distinct region of the space delineated by these three efforts.
We adopt the heterogeneous-event data setting of EBES but target a \textbf{forecasting} task rather than sequence-level classification.
Within the forecasting setting, we complement EasyTPP and HoTPP in two ways.
First, we lift the MTPP restriction, so events may carry heterogeneous features.
Second, we evaluate both \emph{prediction accuracy} and \emph{distributional realism} and use their combination to probe the trade-off between the two.
 
\subsection{Metrics for Generative Sequence Models}
\label{sec:rw-metrics}

We divide metrics for generative event-sequence models into two groups.
% \emph{Distributional realism} metrics measure how close the distribution of generated continuations is to the distribution of true ones.

\paragraph{Prediction accuracy metrics.}
They measure how close, on average, the forecast $\hat S_{H+1:H+L}$ is to the ground-truth continuation $S_{H+1:H+L}$.
The traditional TPP metrics measure how well the model predicts a single step ahead.
The classic examples are time MAE or RMSE, mark accuracy, and next-event negative log-likelihood~\citep{easytpp}.
They are straightforward to compute but are known to correlate only weakly with long-horizon forecasting quality~\citep{hotpp}.
 
Optimal Transport Distance (OTD) for event sequences aligns predicted and ground-truth events via a minimum-cost edit over insertions, deletions, and event moves, with per-operation costs chosen by hand.
It is widely used in TPP and MTPP evaluation~\citep{def, cdiff} and handles length mismatches, but its method ranking is sensitive to the cost hyperparameters~\citep{yang2022transformer}.
Temporal mean Average Precision (T-mAP) instead aligns events by a one-to-one matching over a time tolerance and averages the resulting per-label average precision, replacing the per-operation costs with a single tolerance hyperparameter~\citep{hotpp}.
Neither extends to events with numerical side features.
Our TMS (\S\ref{sec:metrics}) is motivated by the same alignment intuition and extends it: it replaces the hand-chosen costs with data-driven normalization and scores an arbitrary number of numerical and categorical features on a common scale.
 
\paragraph{Distributional realism metrics.}
They measure how closely the distribution of generated continuations matches the distribution of true ones.
The classifier two-sample test (C2ST)~\citep{c2st} measures distributional similarity of generated samples by training a discriminator to separate real from generated samples; the resulting accuracy serves as a proxy for distributional distance.
The idea applies to any data types and underlies our cDFS metric (\S\ref{sec:metrics}) with necessary modifications to support heterogeneous types and conditioning on history.
 
TabSyn~\citep{tabsyn} popularized \emph{Shape} (Kolmogorov--Smirnov distance per feature) and \emph{Trend} (closeness of pairwise feature dependencies) scores, which have since been used by follow-up tabular-diffusion work.
These scores audit the marginal and low-order joint structure of generated rows but ignore the temporal structure.
We include them as \uline{auxiliary distributional metrics for context} in Appendix~\ref{app:more-res:metric-corr}.

\begin{semitransparentsectionbox}
None of these metrics is fully suitable for assessing forecasting accuracy or distributional realism on heterogeneous event sequences.
We bridge this gap by introducing in \S\ref{sec:metrics} two novel metrics: TMS for prediction accuracy, and cDFS for distributional realism.    
\end{semitransparentsectionbox}

%-----------------------------------------------------------
\section{Models for Evaluation}
\label{sec:methods}
%-----------------------------------------------------------

We choose representative methods, one from each major class of generative approaches relevant to event sequences: a latent diffusion model (our tailored \textbf{LDM}, \S\ref{sec:ldm}), a data diffusion model (\textbf{DDM}, \S\ref{sec:cdiff}), and a SoTA non-diffusion method (\textbf{DEF}, \S\ref{sec:def}).
We adapted them for general event sequences (see \uline{implementation details} in Appendix~\ref{app:methods}).
For completeness, we include GPT-style methods (\S\ref{sec:methods:gpt}) and statistical baselines (\S\ref{sec:methods:baselines}) in the comparison.

\subsection{Baselines}\label{sec:methods:baselines}
 
We include three simple methods.
\textbf{Repeat} returns the last $L$ events of the history with their inter-arrival times.
\textbf{HistSampler} draws $L$ events from the history uniformly at random with replacement.
Baselines of this kind are rarely reported in the event-sequence literature; HoTPP~\citep{hotpp} is a notable exception for classical MTPP, but we did not encounter \textbf{Repeat} in the literature, although it performs competitively well (see \S\ref{sec:results}).
\textbf{Mode} sets each categorical dimension to its historical mode and each numerical dimension to the historical median; the time channel uses the median historical inter-arrival time.
 
\subsection{GPT-based}\label{sec:methods:gpt}
Because event sequences are fundamentally sequential, we borrow architectural ideas from large language model (LLM) practice rather than confining ourselves to classical TPP designs.
We instantiate a GPT-2-style decoder in two regimes.
\textbf{AR} is autoregressive: conditioned on the prefix history, it factorizes forecasting through successive next-event predictions.
\textbf{MT} uses the same backbone but is \emph{not} autoregressive: it maps the history to $L$ parallel predictions of future events in a single forward pass (no causal chain over future steps), trained with the sum of per-event negative log-likelihoods across those $L$ slots.

\subsection{Latent Diffusion (LDM)}\label{sec:ldm}

Our latent diffusion method adapts TabSyn~\citep{tabsyn} to the event-sequence setting.
We first train a $\beta$-VAE~\citep{betavae} to encode and decode the features of a single event, treating the time stamp as an additional feature alongside the numerical and categorical ones.
We then train a Gaussian diffusion model~\citep{karras2022elucidating} on sequences of latent event representations, treating a window of $L$ events as a single vector in $\bbR^{L d}$, where $d$ is the per-event latent dimensionality.
To sample $L$ events, we draw a noise vector in $\bbR^{L d}$, run the reverse diffusion process, split the resulting vector into $L$ chunks, and decode each with the VAE decoder.
 
We implement forecasting through conditional generation.
A history embedding produced by an RNN over the past events is used as the primary conditioning signal for the diffusion model.
This embedding summarizes long-range history but may lose information about the most recent events.
To preserve recency, we also feed the latent codes of the last $L$ events in the history as a ``raw'' context.
% Using a combination of compressed history and raw recent events is common in diffusion-based TPP forecasting~\citep{tcddpm, addthin, cdiff, nonar}.
The overall scheme is shown in Figure~\ref{fig:ldm}.

Our LDM is structurally close to the ADiff4TPP method proposed in~\cite{adiff4tpp}.
We compare them \uline{structurally} (Appendix~\ref{app:methods:ldm-vs-adiff4tpp}) and \uline{empirically} (Appendix~\ref{app:more-res:adiff4tpp}).
 
\subsection{Data-space Diffusion Model (DDM)}\label{sec:cdiff}

We use CDiff~\citep{cdiff} as the basis for our representative data-space diffusion method.
CDiff was originally proposed for MTPP forecasting and combines Gaussian diffusion over inter-arrival times with uniform categorical diffusion over event marks~\citep{hoogeboom2021argmax}, training the model to predict multiple future events simultaneously.
The method extends naturally to general event sequences with minimal modification: we retain Gaussian diffusion for inter-arrival times and for numerical features, and use uniform categorical diffusion for each categorical feature.
The method is illustrated in Figure~\ref{fig:ddm}.

We apply classifier-free guidance as in our LDM (Appendix~\ref{app:shared}); the \uline{remaining implementation choices} follow Appendix~\ref{app:methods}.

\subsection{DEF}\label{sec:def}

DEF~\citep{def} is a state-of-the-art method for long-horizon forecasting of MTPP.
The method is built on a backbone model that generates event candidates.
The authors introduced a Horizon Matching Loss that finds the optimal matching between ground truth and event candidates.
The backbone is trained to generate more candidates than typically occur within the horizon, so some candidates remain unmatched.
Each candidate also receives a presence probability, trained jointly to predict whether it will be matched.
At inference time, these probabilities select the events from the candidate pool to form the prediction.
For the schematic illustration of the method, refer to the Figure~\ref{fig:def}.
We adapt DEF to general event sequences by predicting the full heterogeneous feature vector at each candidate event instead of a time and a single mark, and, since our setting fixes the horizon to $L$ events, we select the top-$L$ candidates by presence score rather than by a time cutoff (Appendix~\ref{app:methods}).

%-----------------------------------------------------------
\section{Benchmark Description}\label{sec:benchmark}
%-----------------------------------------------------------

All methods, metrics, and helpers are implemented in PyTorch, in a single open-source repo\footnote{\url{https://github.com/xaosina/event-sequence-forecasting}}.

\subsection{Datasets}\label{sec:datasets}

\begin{table}[h]
\centering
\caption{%
    Dataset statistics. Number of sequences in each split, average sequence length, number of categorical and numerical features describing each event, and overall time span.
}
\label{tab:datasets}
\begin{tabular}{lccccc}
\toprule
Dataset & \# seq. (train / test) & avg. seq. len. & \# num. (incl. time) & \# cat. & time span \\
\midrule
Gender & 11K / 2.8K & $404.31 \pm 292.48$ & 2 & 2 & 15 months \\
Age & 24K / 6K & $881.00 \pm 124.75$ & 2 & 2 & 2 years \\
Alphabattle & 100K / 35K & $366.91 \pm 212.58$ & 2 & 4 & 2 years \\
\bottomrule
\end{tabular}
\end{table}

After a careful review of the available public corpora, we found three that meet our criteria:
\textbf{Gender}\footnote{\url{https://www.kaggle.com/competitions/python-and-analyze-data-final-project/data}}, \textbf{Age}\footnote{\url{https://ods.ai/competitions/sberbank-sirius-lesson}}, and \textbf{Alphabattle}\footnote{\url{https://www.kaggle.com/datasets/mrmorj/alfabattle-20}}.
All three are card transaction datasets with long enough sequences, enough users, and no data artifacts.
Each event includes at least a timestamp, a transaction amount, and an MCC code.
The key statistics are in Table~\ref{tab:datasets}.
Notably, no non-financial public corpus met the size, length, and structure criteria simultaneously. This is a limitation of the current public data landscape rather than of the benchmark design: although the forecasting problem itself spans many domains, the data behind it is produced in privacy-sensitive industries such as banking and healthcare, and rich corpora are rarely released publicly.
The \uline{corpora we rejected and the reasons why} are listed in Appendix~\ref{app:datasets-rejected}; \uline{further preprocessing and splitting details} are provided in Appendix~\ref{app:preprocessing}.

To probe transferability beyond transactions, we additionally evaluate the benchmark on \uline{two TPP corpora (Retweet, Amazon)} in Appendix~\ref{app:tpp}; both fail our selection criteria, having far fewer and much shorter sequences, and each event carries only a timestamp and a categorical mark. Despite this, the accuracy--realism trade-off and the leading models on each axis carry over.

\subsection{Metrics}\label{sec:metrics}
No existing metric covers our setting (\S\ref{sec:related}): accuracy metrics look one step ahead or, like OTD and T-mAP, leave numerical features aside, and realism metrics ignore the history. We therefore introduce a pair of metrics for an arbitrary number of heterogeneous features.

The \emph{Temporal Matching Score} (TMS) measures \emph{accuracy} --- how closely, event by event, the prediction matches the ground-truth continuation.
The \emph{Conditional Detection Fooling Score} (cDFS) measures \emph{distributional realism} --- how difficult it is for a learned discriminator to tell generated continuations apart from real ones given the same history.
As auxiliary per-feature distributional diagnostics, we also describe (Appendix~\ref{app:realism:shape-trend}) and report (Appendix~\ref{app:more-res:metric-corr}) the \uline{Shape and Trend scores} from~\citet{tabsyn}; the \uline{correlation of TMS with matched per-feature submetrics} is in Appendix~\ref{app:more-res:tms-submetrics}.

\paragraph{Conditional Detection Fooling Score (cDFS).}
cDFS is a conditional variant of the classifier two-sample test~\citep{c2st}.
For each test example, we form two sequences sharing the same history $S_{1:H}$: the true continuation $S_{H+1:H+L}$ and a model-generated continuation $\hat S_{H+1:H+L}$.
We train a discriminator to predict, from the generated or true continuation, optionally prepended with the last $H'$ events of history, whether it is real or generated.
The metric is defined as
\begin{equation*}
    \mathrm{cDFS} \defeq 2 \cdot (1 - \mathrm{AUROC}),
\end{equation*}
where AUROC is the ROC-AUC of the discriminator on a held-out portion of the real/generated pairs.
Clipping AUROC from below at $0.5$, cDFS is bounded in $[0, 1]$: it equals $0$ when the discriminator separates real from generated perfectly and $1$ when it does no better than chance, as expected when the conditional distribution of generated continuations matches that of real continuations.
Absolute values depend on the discriminator, so we interpret the relative ranking.
Varying the discriminator's architecture and capacity confirms that \uline{the ranking is stable} (Spearman $\rho \geq 0.91$; Appendix~\ref{app:realism:detector-ablation}).
Unlike classical C2ST, our discriminator sees the history, so it detects continuations that are plausible in isolation but inconsistent with the preceding events.
We describe \uline{technical details of cDFS computation} in Appendix~\ref{app:realism:cdfs}.
% cDFS is computed with a fixed discriminator architecture; .

\paragraph{Temporal Matching Score (TMS).}
We propose TMS to measure prediction accuracy for event sequences.
It is a dataset-level score inspired by the Optimal Transport Distance (OTD)~\citep{def, cdiff}.
OTD aligns predicted and true events via a minimum-cost edit over insertions, deletions, and moves, but its hand-chosen edit costs~\citep{yang2022transformer} do not extend naturally to events with heterogeneous numerical and categorical features.
TMS uses per-feature, baseline-rescaled normalization and a one-to-one permutation of horizon positions instead of hand-tuned costs and edit alignment.

For each numerical feature $m$ and for time, the paired score $R_1^m$ (resp.\ $R_1^t$) is one minus the mean absolute prediction error divided by the mean absolute error of a horizon-naive baseline built from the history.
It can be viewed as a skill-score variant of the Mean Absolute Scaled Error~\citep{hyndman2006another}: $1$ is perfect, $0$ ties the baseline, and lower values are worse.
For each categorical feature $n$, the paired score $F_1^n \in [0,1]$ is the per-pair macro-$F_1$ between predicted and true horizon labels.
The MASE-style baseline rescaling puts heterogeneous features on a common scale.

To separate ``wrong content'' from ``right content in the wrong order,'' each per-feature score is upgraded to a $\sigma$-parameterized version under a per-pair permutation $\sigma$ between predicted and true horizon positions, with the paired score recovered at $\sigma = \mathrm{id}$.
TMS averages these scores under a single shared $\sigma$ optimized jointly over all features,
\begin{equation}\label{eq:tms}
    \mathrm{TMS}(\mathcal{T}) \defeq \max_{\sigma}\, \frac{1}{1 + M + N}\!\left( R_1^t(\sigma) + \sum_{m=1}^M R_1^m(\sigma) + \sum_{n=1}^N F_1^n(\sigma) \right).
\end{equation}
The maximum decomposes per pair into a Hungarian assignment~\citep{kuhn1955hungarian} on an $L \times L$ feature-averaged cost matrix.
Because time is one of the matched features, reordering events still incurs a penalty when it worsens temporal alignment.
Because the same $\sigma$ enters every term, TMS is a single matched score rather than an average of independently aligned per-feature diagnostics.
\uline{Full definitions} of the per-feature scores, the present-class normalization for $F_1^n$, the proof that the per-feature numerators decompose into per-pair contributions, and matched per-feature variants are deferred to Appendix~\ref{app:tms}.

%-----------------------------------------------------------
\section{Wasserstein BaryBooster}\label{sec:barybooster}
%-----------------------------------------------------------

What distinguishes a generative forecaster is that it predicts many possible future trajectories rather than one, and for a good model these trajectories follow the distribution of the real data. Accuracy, however, is awarded to a single continuation, so this potential is left unused. To harvest it, we introduce \emph{Wasserstein BaryBooster}, a post-hoc procedure that turns $N$ stochastic samples from any base generator into a single continuation optimal under TMS.

\begin{semitransparentsectionbox}
Given a history $S$, we draw $\hat S_1, \ldots, \hat S_N$ from the base model and return
\begin{equation}
    \hat S^\star \;\defeq\; \arg\min_{\hat S}\; \sum_{n=1}^{N} d_{\mathrm{TMS}}\!\left(\hat S,\, \hat S_n\right),
    \label{eq:barybooster}
\end{equation}
where $d_{\mathrm{TMS}}(A, B) = \min_{\sigma} \sum_{i \in \mathcal{H}} \bar C_{(A, B)}(i, \sigma(i))$ is the TMS transport cost between two continuations under the feature-averaged cost matrix $\bar C$~(Appendix~\ref{app:tms:hungarian}).
 \end{semitransparentsectionbox}

Equation~\eqref{eq:barybooster} is the discrete TMS analog of a Wasserstein barycenter~\citep{agueh2011barycenters}, which motivates the name; the \emph{booster} suffix reflects that the procedure boosts the predictive accuracy of any underlying generator without retraining.
Note that the procedure is not limited to TMS: any other metric can take its place in Eq.~\eqref{eq:barybooster}.
We use $N = 30$; the \uline{optimization procedure} is given in Appendix~\ref{app:barybooster}.

\section{Experiments}\label{sec:results}

\begin{table}[ht]
\centering
\caption{%
TMS and cDFS at horizon $L=32$. 
Within the baseline group, we highlight \textbf{the best}, \uline{the second best}, and \dashuline{the third best} methods.
BaryBoosted methods are highlighted when they would rank first/second/third after being added to the pool. \colorbox{red!15}{Red} cells mark cDFS $<0.07$ (easily detected).
}
\label{tab:metrics_h32}
\begin{tabular}{lcccccc}
\toprule
Method & \multicolumn{3}{c}{TMS ↑} & \multicolumn{3}{c}{cDFS ↑} \\
\cmidrule(lr){2-4} \cmidrule(lr){5-7}
 & Alphabattle & Gender & Age & Alphabattle & Gender & Age \\
\midrule
Hist. & 0.382 & 0.221 & 0.366 & \uline{0.544} & \uline{0.476} & \uline{0.234} \\
Mode & 0.203 & 0.047 & 0.263 & \cellcolor{red!15}0.000 & \cellcolor{red!15}0.000 & \cellcolor{red!15}0.000 \\
Rep. & \textbf{0.440} & \uline{0.361} & \uline{0.454} & \cellcolor{red!15}\dashuline{0.009} & \cellcolor{red!15}\dashuline{0.050} & \cellcolor{red!15}0.065 \\
AR & 0.286 & 0.188 & 0.354 & \cellcolor{red!15}0.000 & \cellcolor{red!15}0.002 & \cellcolor{red!15}0.000 \\
MT & 0.286 & 0.153 & 0.360 & \cellcolor{red!15}0.000 & \cellcolor{red!15}0.000 & \cellcolor{red!15}0.000 \\
LDM & \dashuline{0.405} & \dashuline{0.318} & 0.426 & \textbf{0.856} & \textbf{0.907} & \textbf{1.000} \\
DDM & 0.251 & 0.107 & \dashuline{0.441} & \cellcolor{red!15}0.004 & \cellcolor{red!15}0.002 & \dashuline{0.202} \\
DEF & \uline{0.424} & \textbf{0.413} & \textbf{0.508} & \cellcolor{red!15}0.000 & \cellcolor{red!15}0.002 & \cellcolor{red!15}0.000 \\
\midrule
LDM-WB (ours)& \textbf{0.495} & \textbf{0.436} & \textbf{0.522} & \dashuline{0.177} & \dashuline{0.256} & \dashuline{0.224} \\
DDM-WB (ours)& 0.274 & 0.136 & \dashuline{0.480} & \cellcolor{red!15}0.001 & \cellcolor{red!15}0.001 & 0.102 \\
\bottomrule
\end{tabular}
\end{table}

\subsection{Evaluation Protocol}\label{sec:eval}

We forecast at fixed horizons $L \in \{4, 16, 32, 64\}$.
Main results use $L{=}32$ (Table~\ref{tab:metrics_h32}); see \uline{$L\in\{4,16,64\}$} in Appendix~\ref{app:more-res:horizon}, where the key conclusions hold.
We report both cDFS and TMS (\S\ref{sec:metrics}) so comparisons reflect both distributional realism and prediction accuracy, not just a single score.
We also evaluate \emph{cross-horizon} models: those trained at shorter $L$ are semi-autoregressively rolled out to longer $L$ (e.g., LDM trained on $L=16$ and evaluated on a longer horizon is denoted \textbf{LDM/16}), to compare this approach to direct one-shot forecasting.
Additionally, we apply \emph{Wasserstein BaryBooster} (\S\ref{sec:barybooster}) to pretrained diffusion models and mark them with the \textbf{-WB} suffix.
Details on \uline{data sampling and splitting} are in Appendix~\ref{app:preprocessing}.
We report \uline{run time and GPU memory usage} in Appendix~\ref{app:runtime-h64}.
We address \uline{statistical significance and metric stability across runs} in Appendix~\ref{app:more-res:seeds}.

\subsection{Results}

\paragraph{Performance overview.}
Table~\ref{tab:metrics_h32} reveals a sharp split between the two metrics.
LDM is the only generative model to achieve a non-trivial cDFS, except for DDM on Age.
Every remaining learned model, including the state-of-the-art DEF, collapses to cDFS values near zero, indicating that a simple discriminator separates their continuations from real ones with near-perfect accuracy. Appendix~\ref{app:more-res:qualitative} shows \uline{generated samples} for a single client.
Among the remaining methods, HistSampler is the closest competitor to LDM, but it produces less realistic sequences  because it ignores temporal dependencies.
Table~\ref{tab:metrics_h32} reports boosted variants denoted \textbf{LDM-WB} and \textbf{DDM-WB}; applying the same procedure to other generators yields little or no comparable improvement, so we omit those boosted variants from the main tables and report them in Appendix~\ref{app:barybooster-other}.

The prediction accuracy ranking tells a different story.
DEF is the best or second-best method across all datasets, confirming its position as the strongest non-diffusion forecaster.
Repeat is the strongest baseline on Alphabattle and second-best on the other two datasets, beating every learned method on Alphabattle and outperforming both LDM and DDM almost everywhere.
AR and MT remain below HistSampler on both cDFS and TMS.
Their numerical features collapse to the conditional median, losing nearly all variance --- the pathology of Figure~\ref{fig:intro-example} (Appendix~\ref{app:more-res:qualitative}).
Cross-horizon models are almost never better than direct one-shot models trained at the target horizon.

\begin{figure}[ht]
    \centering
    \includegraphics[width=\linewidth]{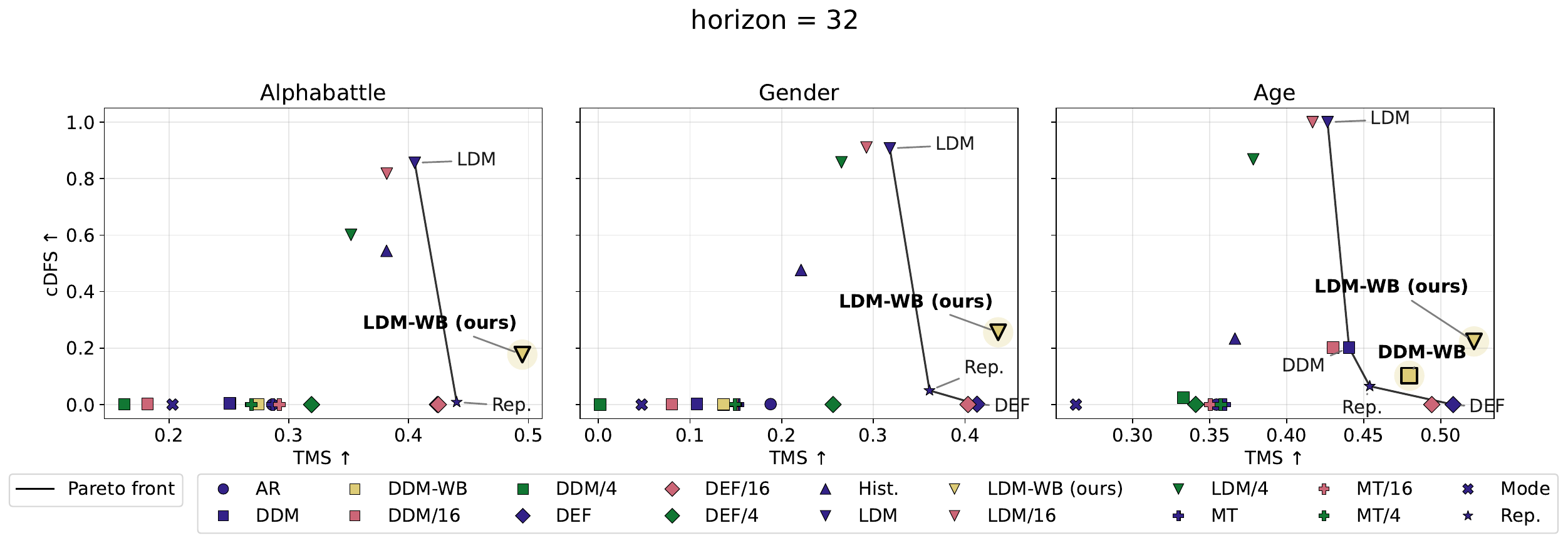}
    \caption{%
        Trade-off between TMS and cDFS for each dataset on horizon $L=32$.
        The solid line shows the Pareto front over baseline methods.
        }
    \label{fig:pareto_32}
\end{figure}

\paragraph{Accuracy--realism trade-off.}
Figure~\ref{fig:pareto_32} shows the trade-off explicitly.
Methods with high prediction accuracy show low distributional realism and vice versa.
Interestingly, Repeat lies on the Pareto front of the baseline pool across all datasets, despite being a one-line, non-parametric procedure. 
This highlights the room for specialized generative models that surpass Repeat's accuracy without sacrificing realism.
% The trade-off is not merely a definitional consequence of point-wise vs. distributional metrics: a method that simply collapsed to the conditional mode would lose on both axes (see Mode baseline, Table~\ref{tab:metrics_h32}), and conversely, a perfectly-calibrated sampler would score well on both.
% The observed front, therefore, reflects modeling choices, not the metrics' geometry.

\paragraph{BaryBoosting diffusion models.}

Wasserstein BaryBooster (\S\ref{sec:barybooster}) realizes the predictive potential of the LDM: when applied, it produces the highest TMS across all datasets while still retaining a non-trivial cDFS (see Table~\ref{tab:metrics_h32}).
Applying the same procedure to DDM yields only modest TMS gains, reflecting DDM's weaker base generations.
Boosting helps most for diffusion-based generators (Appendix~\ref{app:barybooster-other}): the gain is largest for diffusion models, absent for DEF, and small for the remaining non-diffusion methods.

Figure~\ref{fig:pareto_32} clearly shows that BaryBoosting moves the Pareto front: by aggregating multiple LDM samples around their TMS barycenter, it converts part of the parent model's realism into the highest TMS.
This converts a generative model into a competitive predictor while leaving the original LDM available whenever distributional realism matters more than prediction accuracy.
The effect is strongest at $L=4$ (Figure~\ref{fig:pareto_4}), where boosting shifts LDM from near-bottom TMS to best-in-class predictor.

The gain is not confined to the optimized cost: Table~\ref{tab:wb-ind-main} shows that BaryBoosting LDM also improves many other prediction metrics that BaryBooster does not optimize directly.
Appendix~\ref{app:more-res:wb-ind-eval} extends this \uline{independent evaluation} to all methods; DDM also improves after boosting, with a few exceptions.

\begin{table}[ht]
\centering
\caption{BaryBoosting improves prediction metrics beyond the optimized TMS ($L=32$; excerpt of Table~\ref{tab:more_metrics_h32}).}
\label{tab:wb-ind-main}
\resizebox{\textwidth}{!}{%
\begin{tabular}{lccccccccc}
\toprule
Method & \multicolumn{3}{c}{R1 (amount) ↑} & \multicolumn{3}{c}{R1 (time) ↑} & \multicolumn{3}{c}{F1 (mcc) ↑} \\
\cmidrule(lr){2-4} \cmidrule(lr){5-7} \cmidrule(lr){8-10}
 & Alphabattle & Gender & Age & Alphabattle & Gender & Age & Alphabattle & Gender & Age \\
\midrule
LDM & 0.479 & 0.180 & 0.216 & 0.148 & 0.144 & 0.074 & 0.400 & 0.364 & 0.323 \\
LDM-WB (ours) & \textbf{0.583} & \textbf{0.334} & \textbf{0.438} & \textbf{0.337} & \textbf{0.331} & \textbf{0.331} & \textbf{0.459} & \textbf{0.448} & \textbf{0.371} \\
\bottomrule
\end{tabular}
}
\end{table}

\section{Discussion}\label{sec:discussion}

\paragraph{Outlook}
A realistic generative model captures the whole distribution of possible futures, mirroring the real world, where the same history can lead to many different futures.
We show how this property can be put to use: aggregating samples with the BaryBooster yields a state-of-the-art point forecast with no additional training for the target metric.
This opens the way to foundation models for event sequences: a single generative model, trained once, from which many downstream tasks can be solved.

\paragraph{Limitations}\label{sec:discussion:limitations}
Our empirical study uses a relatively small number of datasets and focuses narrowly on card transactions.
Although we made a best-effort attempt to diversify the benchmark, practical constraints on data quality and schema compatibility limited the final pool; details on \uline{considered but excluded sources} are reported in Appendix~\ref{app:datasets-rejected}.
In addition, the Wasserstein BaryBooster introduces substantial computational overhead, since it requires generating multiple candidate trajectories and solving an iterative barycentric aggregation procedure at inference time.
While this improves predictive performance, the extra compute may be prohibitive in latency-sensitive or resource-constrained deployments.
The overhead is, however, an inference-time choice: $N=5$ already retains ${\approx}75\%$ of the TMS gain at one sixth of the sampling cost (Appendix~\ref{app:barybooster-n}), and the $N$ samplings parallelize across GPUs.

\paragraph{Broader impacts}\label{sec:discussion:impacts}
The forecasts studied in this paper can support planning and risk analysis. Since a generative model produces many plausible continuations, analysts can explore the different paths a client's future may take, rather than a single average forecast. Our metrics, benchmark, and methods are not specific to transactions and can bring these benefits to event sequences in other domains, such as healthcare and e-commerce. On the other hand, realistic synthetic transactions could be misused, e.g., to evade fraud detection; in this work, all models are trained on public, anonymized data.

\begin{ack}
The work was supported by the grant for research centers in the field of AI provided by the Ministry of Economic Development of the Russian Federation in accordance with the agreement 000000C313925P4F0002 and the agreement No.~139-10-2025-033.
The authors declare no competing interests.
\end{ack}

\bibliographystyle{unsrtnat}
\bibliography{biblio}

\appendix

\etocdepthtag.toc{appendix}
\etocsettagdepth{main}{none}
\etocsettagdepth{appendix}{subsection}
\etocsettocstyle{\section*{Appendix Contents}}{}
\tableofcontents

\section{Generative Models for Event Data: Extended Discussion}\label{app:rw-models}

This appendix expands the discussion of the related generative models summarized in Table~\ref{tab:methods} and referenced in \S\ref{sec:rw-models}, grouped by the three lines of work they belong to.

\paragraph{Tabular data.}
Event sequences can be viewed as time-ordered sequences of tabular rows, so diffusion models for heterogeneous tabular data are directly relevant. TabSyn~\citep{tabsyn} is a latent diffusion model in which a $\beta$-VAE~\citep{betavae} encodes mixed-type features into a latent space, and Gaussian diffusion~\citep{karras2022elucidating} generates new latents that are decoded back to rows. TabDiff~\citep{tabdiff} is the data-diffusion counterpart: it couples masked discrete diffusion~\citep{shi2024simplified, mdlm} for categorical features and Gaussian diffusion~\citep{karras2022elucidating} for numerical features, generating rows directly. Neither model, however, natively targets sequences.

\paragraph{Temporal point processes.}
Temporal point processes (TPP) are the classical formalism for random event times, and most prior generative work on event sequences falls under TPP. In \emph{unmarked} TPP, each event is characterized by its time only. Add-and-Thin~\citep{addthin} introduces a diffusion-inspired noising and denoising process tailored for point processes, and its successor EdiTPP~\citep{editpp} casts sequence generation as edit-based flow matching~\citep{editflow}. \emph{Spatio-temporal} point processes (STPP) extend events with numerical coordinates; Point Set Diffusion~\citep{psd} generalizes Add-and-Thin to this setting. \emph{Marked} TPP (MTPP) adds a single categorical mark per event and is the most studied variant. ADiff4TPP~\citep{adiff4tpp} uses a $\beta$-VAE to embed (time, mark) pairs and applies flow matching with non-uniform noise scales over the sequence. CDiff~\citep{cdiff} combines uniform categorical diffusion~\citep{hoogeboom2021argmax} for marks with Gaussian diffusion for inter-arrival times and is trained to predict multiple future events jointly. DEF~\citep{def} is a recent non-diffusion method that predicts a fixed set of candidate future events and scores them, keeping the top-scoring ones.

\paragraph{General event sequences.}
The MTPP methods above naturally extend to heterogeneous events: ADiff4TPP~\citep{adiff4tpp}, CDiff~\citep{cdiff}, and DEF~\citep{def} all generalize to additional numerical and categorical features. TabDiT~\citep{tabdit} is a latent-diffusion approach closely related to ADiff4TPP, but it targets attribute-conditional generation rather than history-conditional (forecasting). UniTTab~\citep{unittab} is a transformer model tailored for heterogeneous event sequences, but its authors do not study generative modeling --- they use the transformer only to extract sequence representations for downstream discriminative tasks.

\section{Benchmarks for Event Sequences: Extended Discussion}\label{app:rw-benchmarks}

This appendix expands the discussion from \S\ref{sec:rw-benchmarks} of the three prior benchmarks closest to our work --- EasyTPP, HoTPP, and EBES --- and details why each falls short of the heterogeneous, realism-aware forecasting setting we target.

\paragraph{EasyTPP.}
EasyTPP~\citep{easytpp} is the first open benchmarking effort for neural TPPs. It provides a unified interface to several widely used TPP datasets, along with reference implementations of popular neural TPP models. Evaluation covers both next-event prediction, reported as time mean absolute error and mark classification accuracy, and long-horizon prediction via optimal transport distance on multi-step forecasts. Two properties of the benchmark limit its scope relative to our setting. First, events are restricted to a single categorical mark, i.e., EasyTPP operates strictly within the MTPP formalism --- there is no facility for additional numerical or categorical features, which rules out the domains (finance, retail) where heterogeneous events are the norm. Second, all reported metrics quantify closeness of the prediction to the ground truth; the benchmark does not assess whether generated sequences are distributionally realistic, and in particular provides no way to detect mode collapse or distributional mismatch.

\paragraph{HoTPP.}
HoTPP~\citep{hotpp} explicitly targets long-horizon forecasting within the MTPP formalism. Its authors point out that next-event metrics correlate poorly with long-horizon quality, and that the classical optimal-transport distance for multi-step evaluation is sensitive to its hand-chosen hyperparameters. As a remedy, they propose Temporal mean Average Precision (T-mAP), a metric inspired by object detection that evaluates predicted event-label distributions over the full horizon. It also reports a single accuracy-style metric per configuration, without a dedicated account of distributional realism.

\paragraph{EBES.}
EBES~\citep{ebes} is the closest prior benchmark in terms of data structure: it explicitly targets heterogeneous event sequences with irregular time stamps and a mix of numerical and categorical features, assembles a large collection of real-world datasets, and implements representative sequence models. Its scope, however, is \emph{discriminative}: EBES evaluates sequence-level classification, where each sequence carries a single target label. It does not address forecasting or any form of generative modeling.

\section{Temporal Matching Score: full definition}\label{app:tms}

This appendix gives the full construction summarized in equation~\eqref{eq:tms}.
We define the per-feature paired scores, their permutation-dependent versions, the matched per-feature variants, and the resulting TMS, and we derive the per-pair decomposition that makes optimization a Hungarian assignment.

\subsection{Notation}\label{app:tms:notation}
Let $\mathcal{T} = \{(S, \hat S)\}$ denote the evaluation set of sequence-prediction pairs, all sharing forecast-horizon length $L$, and let $\mathcal{H} = \{H{+}1,\ldots,H{+}L\}$ be the horizon index set.
Let $\Delta t_i = t_i - t_{i-1}$, let $x_i^m$ be the $m$-th of $M$ numerical features of the $i$-th event, and let $k_i^n$ be the $n$-th of $N$ categorical features.
Hats denote predicted quantities, $\med{\cdot}$ denotes the median, and $\bbE_{(S, \hat S)}[\cdot]$ is the uniform average over $\mathcal{T}$.

\subsection{Paired \texorpdfstring{$R_1$}{R1} for numerical features and time}\label{app:tms:r1}
Paired $R_1$ answers: at each horizon step, how close are predicted magnitudes and timestamps to the ground truth, relative to a naive baseline?
It compares mean absolute error to a simple baseline built only from the past (constant history median for features, uniform spacing from history for time), so the score is a unitless skill measure: higher means the model beats that naive reference.

The paired score for a numerical feature $m$ and for time is the mean absolute prediction error rescaled by the mean absolute error of a horizon-naive forecast built from the history,
\begin{equation*}
    R_1^m \defeq 1 -
    \frac{
        \bbE_{(S, \hat S)}\!\left[\sum_{i \in \mathcal{H}} |\hat x_i^m - x_i^m| \right]
    }{
        \bbE_{(S, \hat S)}\!\left[\sum_{i \in \mathcal{H}} |x_i^m - \med{x_{1:H}^m}| \right]
    },
\end{equation*}
\begin{equation*}
    R_1^t \defeq 1 -
    \frac{
        \bbE_{(S, \hat S)}\!\left[\sum_{i \in \mathcal{H}} |\hat t_i - t_i| \right]
    }{
        \bbE_{(S, \hat S)}\!\left[\sum_{i \in \mathcal{H}} \left| t_i - \bigl(t_H + (i-H)\med{\Delta t_{1:H}}\bigr) \right| \right]
    }.
\end{equation*}
The numerical baseline is the constant history median; the time baseline is a uniform-spacing forecast from the last historical event using the median historical inter-event gap.
This closely resembles the Mean Absolute Scaled Error of~\citet{hyndman2006another}, with a horizon-naive baseline in place of the in-sample one-step-ahead naive forecast, expressed as a skill score: $R_1 = 1$ corresponds to zero prediction error, $R_1 = 0$ means the model's average error equals the baseline's, and $R_1 < 0$ means the model is worse than the baseline on average --- with no lower bound.
Heterogeneous features become commensurable through this baseline rescaling, so $R_1$ values can be averaged across features that live on incomparable units.

\subsection{Paired \texorpdfstring{$F_1$}{F1} for categorical features}\label{app:tms:f1}
Paired $F_1$ answers: when we compare predicted and true labels at the same horizon index, how well does the model recover each class, balancing false positives and false negatives?
It is macro-$F_1$ over classes that actually appear in the pair, so rare or absent labels do not dominate the average.

For a categorical feature $n$, let
\begin{equation*}
    \mathrm{TP}_c^n \defeq \sum_{i \in \mathcal{H}} \mathbf{1}[\hat k_i^n = c]\,\mathbf{1}[k_i^n = c],
\quad
    n_{c,\mathrm{pred}}^n \defeq \sum_{i \in \mathcal{H}} \mathbf{1}[\hat k_i^n = c],
\quad
    n_{c,\mathrm{true}}^n \defeq \sum_{i \in \mathcal{H}} \mathbf{1}[k_i^n = c],
\end{equation*}
and let $\mathcal{C}_{(S,\hat S)}^n \defeq \{c : n_{c,\mathrm{pred}}^n + n_{c,\mathrm{true}}^n > 0\}$ be the set of class labels appearing in either continuation.
Using the standard identity $2\, P_c R_c / (P_c + R_c) = 2\,\mathrm{TP}_c / (n_{c,\mathrm{pred}} + n_{c,\mathrm{true}})$, the paired score is the macro-$F_1$ at matched horizon positions, normalized by the number of present classes per pair and averaged over $\mathcal{T}$,
\begin{equation*}
    F_1^n \defeq \bbE_{(S, \hat S)}\!\left[ \frac{1}{|\mathcal{C}_{(S,\hat S)}^n|} \sum_{c \in \mathcal{C}_{(S,\hat S)}^n} \frac{2\, \mathrm{TP}_c^n}{n_{c,\mathrm{pred}}^n + n_{c,\mathrm{true}}^n} \right].
\end{equation*}
The bincounts $n_{c,\mathrm{pred}}^n$, $n_{c,\mathrm{true}}^n$ and the present-class set $\mathcal{C}_{(S,\hat S)}^n$ are invariant under permutations of horizon positions; only $\mathrm{TP}_c^n$ depends on the alignment.
Restricting the average to present classes coincides with standard macro-$F_1$ when all global labels appear, but avoids inflating the normalizer in pairs where most classes are absent --- a common situation for high-cardinality features.
In particular $F_1^n \in [0, 1]$.

\subsection{Permutation-dependent scores}\label{app:tms:sigma}
Strict index-by-index comparison can punish a model that gets the right events but in a permuted order along the horizon.
Permutation-dependent scores re-label which true position is paired with each predicted position: $\sigma$ is a one-to-one alignment between predicted and true horizon slots, and the formulas below evaluate $R_1$ and $F_1$ after applying that alignment.

We parameterize each per-feature score by a per-pair permutation $\sigma_{(S,\hat S)} \colon \mathcal{H} \to \mathcal{H}$ that aligns predicted positions to true positions.
Substituting $x_{\sigma(i)}$, $t_{\sigma(i)}$, and $k_{\sigma(i)}$ for $x_i$, $t_i$, and $k_i$ in the paired definitions gives
\begin{equation*}
    R_1^m(\sigma) \defeq 1 -
    \frac{
        \bbE_{(S, \hat S)}\!\left[\sum_{i \in \mathcal{H}} |\hat x_i^m - x_{\sigma(i)}^m| \right]
    }{
        \bbE_{(S, \hat S)}\!\left[\sum_{i \in \mathcal{H}} |x_i^m - \med{x_{1:H}^m}| \right]
    }
\end{equation*}
\begin{equation*}
    R_1^t(\sigma) \defeq 1 -
    \frac{
        \bbE_{(S, \hat S)}\!\left[\sum_{i \in \mathcal{H}} |\hat t_i - t_{\sigma(i)}| \right]
    }{
        \bbE_{(S, \hat S)}\!\left[\sum_{i \in \mathcal{H}} \left| t_i - \bigl(t_H + (i-H)\med{\Delta t_{1:H}}\bigr) \right| \right]
    }
\end{equation*}
\begin{equation*}
    F_1^n(\sigma) \defeq \bbE_{(S, \hat S)}\!\left[ \frac{1}{|\mathcal{C}_{(S,\hat S)}^n|} \sum_{c \in \mathcal{C}_{(S,\hat S)}^n} \frac{2\, \mathrm{TP}_c^n(\sigma)}{n_{c,\mathrm{pred}}^n + n_{c,\mathrm{true}}^n} \right],
\quad
    \mathrm{TP}_c^n(\sigma) \defeq \sum_{i \in \mathcal{H}} \mathbf{1}[\hat k_i^n = c]\,\mathbf{1}[k_{\sigma(i)}^n = c].
\end{equation*}
The paired scores correspond to $\sigma = \mathrm{id}$.
Only the numerators depend on $\sigma$: the $R_1$ baselines, the per-class bincount denominators of $F_1^n$, and the present-class set $\mathcal{C}_{(S,\hat S)}^n$ are all invariant under permutations of horizon positions.

\subsection{Matched per-feature scores}\label{app:tms:matched}
Matched per-feature scores ask: if we could reorder the ground truth independently for time, for each numerical feature, and for each categorical feature, what is the best alignment for that feature alone?
They are optimistic upper envelopes of the paired scores and isolate how informative each channel is under its own best reordering.

The matched score of each feature is the supremum of its $\sigma$-parameterized form over per-pair permutations,
\begin{equation*}
    R_1^{m,\star} \defeq \max_{\sigma} R_1^m(\sigma),
\quad
    R_1^{t,\star} \defeq \max_{\sigma} R_1^t(\sigma),
\quad
    F_1^{n,\star} \defeq \max_{\sigma} F_1^n(\sigma).
\end{equation*}
Each maximum is attained by independently choosing $\sigma_{(S,\hat S)}$ for every pair, since the dataset-level denominators are $\sigma$-independent and the numerators decompose into sums of per-pair contributions (\ref{app:tms:hungarian}).
By construction every matched score is at least its paired counterpart.

\subsection{Per-pair decomposition and Hungarian optimization}\label{app:tms:hungarian}
For a fixed feature and pair $(S, \hat S)$, every $\sigma$-dependent numerator above is a sum over $i \in \mathcal{H}$ of a term that depends only on the assigned pair $(i, \sigma(i))$.
For $R_1^m(\sigma)$ and $R_1^t(\sigma)$ this is immediate from their absolute-value form.
For $F_1^n(\sigma)$, swapping the two summations and collapsing the indicator on $c$ gives
\begin{equation*}
    \sum_{c \in \mathcal{C}_{(S,\hat S)}^n} \frac{2\, \mathrm{TP}_c^n(\sigma)}{n_{c,\mathrm{pred}}^n + n_{c,\mathrm{true}}^n}
    = \sum_{i \in \mathcal{H}} \mathbf{1}[\hat k_i^n = k_{\sigma(i)}^n]\, \frac{2}{n_{\hat k_i^n,\mathrm{pred}}^n + n_{\hat k_i^n,\mathrm{true}}^n},
\end{equation*}
which is again a per-pair sum: $n_{c,\mathrm{pred}}^n$ and $n_{c,\mathrm{true}}^n$ are $\sigma$-independent, and on each match $\hat k_i^n = k_{\sigma(i)}^n$.
Each per-feature numerator is therefore $\sum_i C_d(i, \sigma(i))$ for a precomputable $L \times L$ cost matrix $C_d$, and the per-pair maximum is obtained exactly by the Hungarian algorithm.

\subsection{TMS}\label{app:tms:tms}
TMS is the headline accuracy metric: it measures how well the full continuation (time, all numerical features, all categorical features) matches the truth when a single reordering of the horizon is chosen for everyone at once.

TMS applies the same matching to the uniform average of all per-feature scores under a single shared permutation,
\begin{equation*}
    \mathrm{TMS}(\mathcal{T}) \defeq \max_{\sigma}\, \frac{1}{1 + M + N}\!\left( R_1^t(\sigma) + \sum_{m=1}^M R_1^m(\sigma) + \sum_{n=1}^N F_1^n(\sigma) \right).
\end{equation*}
The optimum is again attained per-pair, by Hungarian assignment on the feature-averaged cost matrix $\bar C = \tfrac{1}{1+M+N}\sum_d C_d$.
Because the same $\sigma$ enters every term, TMS is a single matched score rather than an average of independently aligned per-feature ones, and in general lies below the average of the matched per-feature scores $R_1^{m,\star}$, $R_1^{t,\star}$, $F_1^{n,\star}$ defined in \ref{app:tms:matched}.

\section{Distributional Realism Metrics}\label{app:realism}

\subsection{Technical Details of cDFS}\label{app:realism:cdfs}

For each test history, we first take the two continuation tables produced by evaluation: \texttt{orig} (real) and \texttt{gen} (generated).
We then build one binary detection dataset by labeling \texttt{orig} with $y{=}0$ and \texttt{gen} with $y{=}1$, and only after that, splitting this combined table into train/validation/test with proportions $0.7/0.15/0.15$.
In all experiments, conditioning equals target length ($H'=L$), so each sequence is truncated to its last $2L$ events ($L$ condition + $L$ target); train is used for fitting, validation for early stopping, and test for reporting metrics.
Features are preprocessed with the same schema as the source dataset (time feature processing, numerical/categorical 
transforms, and optional log transform for amount), then fed into a 2-layer GRU (hidden size 512, dropout 0.1) and classified by a linear head.
We run three seeds and report mean held-out \texttt{MulticlassAUROC}.
The reported cDFS is
\begin{equation*}
    \mathrm{cDFS} = 2(1-\overline{\mathrm{AUROC}}),
\end{equation*}
with higher values indicating better fooling (harder real/generated discrimination given the same history).

\subsection{Robustness of cDFS to the discriminator choice}\label{app:realism:detector-ablation}

cDFS relies on a learned discriminator, so we check how the metric depends
on the discriminator's architecture and capacity. We recompute cDFS under eight
configurations, $\{\text{GRU}, \text{Transformer}\} \times \{\text{128-1L},
\text{512-1L}, \text{512-2L}, \text{1024-3L}\}$ (hidden width--number of layers),
on all three datasets (Tables~\ref{tab:detector-age}--\ref{tab:detector-alphabattle}).
In all 24 dataset--configuration pairs, the ground truth scores ${\approx}1$ and LDM
is the most realistic model. The values of intermediate methods do move (e.g.,
HistSampler on Age: 0.20--0.83). However, the relative ranking stays stable: Spearman's $\rho$ between the rankings under
each configuration and under GRU-512-2L, used in the paper, computed on the reported
values (ties at three decimals), \textbf{ranges from 0.91 to 1.00}; values below 0.95 come
from reorderings among point predictors with cDFS at or below 0.05, except one case.

That case is Repeat on Age under GRU-1024-3L: 0.705 instead of 0.065 ($\rho=0.91$,
or 0.98 without Repeat), high on both axes of the trade-off plot. The detector
identifies Repeat by the verbatim copy of the history
(Appendix~\ref{app:realism:shape-trend}), and this pattern is easy to fit: all other
configurations except Transformer-1024-3L separate Repeat almost perfectly on their
training set (train AUROC $\geq 0.97$). GRU-1024-3L on Age reaches only 0.73 even on
the training set, so the high cDFS of Repeat reflects a detector that failed to fit,
not realistic continuations.
The same holds for Transformer-1024-3L (train AUROC 0.76--0.90 on Repeat), which
underfits on all datasets and raises almost all methods at once. A larger
discriminator is thus not a stronger one.

GRU-512-2L, the configuration we use throughout the paper, catches the repetition
pattern on all three datasets, which makes it both cheaper and more reliable than
the largest configurations.

\begin{table}[ht]
\centering
\resizebox{\textwidth}{!}{%
\begin{tabular}{lcccccccc}
\toprule
Method & \makecell{GRU\\128-1L} & \makecell{GRU\\512-1L} & \makecell{GRU\\512-2L (default)} & \makecell{GRU\\1024-3L} & \makecell{Tr.\\128-1L} & \makecell{Tr.\\512-1L} & \makecell{Tr.\\512-2L} & \makecell{Tr.\\1024-3L} \\
\midrule
GT & 1.000 & 1.000 & 1.000 & 0.996 & 1.000 & 1.000 & 1.000 & 1.000 \\
Hist. & 0.198 & 0.234 & 0.234 & 0.225 & 0.727 & 0.756 & 0.727 & 0.831 \\
Mode & 0.000 & 0.000 & 0.000 & 0.000 & 0.000 & 0.000 & 0.000 & 0.000 \\
Rep. & 0.031 & 0.037 & 0.065 & 0.705 & 0.008 & 0.003 & 0.002 & 0.334 \\
AR & 0.000 & 0.000 & 0.000 & 0.000 & 0.000 & 0.000 & 0.001 & 0.007 \\
MT & 0.000 & 0.000 & 0.000 & 0.000 & 0.000 & 0.000 & 0.000 & 0.069 \\
LDM & 0.991 & 0.975 & 0.966 & 0.976 & 0.996 & 0.993 & 0.982 & 0.998 \\
DDM & 0.176 & 0.203 & 0.202 & 0.219 & 0.470 & 0.521 & 0.494 & 0.709 \\
DEF & 0.000 & 0.000 & 0.000 & 0.000 & 0.000 & 0.000 & 0.000 & 0.036 \\
LDM-WB & 0.187 & 0.209 & 0.224 & 0.235 & 0.148 & 0.166 & 0.097 & 0.448 \\
\midrule
Spearman $\rho$ vs. default & 1.00 & 1.00 & 1.00 & 0.91 & 0.99 & 0.99 & 0.97 & 0.96 \\
\bottomrule
\end{tabular}
}
\caption{\textbf{Age}: cDFS under eight discriminator configurations, horizon $L=32$. Spearman $\rho$ compares the model ranking under each configuration with the ranking under GRU-512-2L.}
\label{tab:detector-age}
\end{table}

\begin{table}[ht]
\centering
\resizebox{\textwidth}{!}{%
\begin{tabular}{lcccccccc}
\toprule
Method & \makecell{GRU\\128-1L} & \makecell{GRU\\512-1L} & \makecell{GRU\\512-2L (default)} & \makecell{GRU\\1024-3L} & \makecell{Tr.\\128-1L} & \makecell{Tr.\\512-1L} & \makecell{Tr.\\512-2L} & \makecell{Tr.\\1024-3L} \\
\midrule
GT & 1.000 & 1.000 & 1.000 & 1.000 & 1.000 & 1.000 & 1.000 & 1.000 \\
Hist. & 0.427 & 0.493 & 0.477 & 0.470 & 0.612 & 0.673 & 0.535 & 0.871 \\
Mode & 0.000 & 0.000 & 0.000 & 0.000 & 0.000 & 0.000 & 0.000 & 0.000 \\
Rep. & 0.177 & 0.067 & 0.050 & 0.054 & 0.007 & 0.008 & 0.001 & 0.197 \\
AR & 0.001 & 0.002 & 0.002 & 0.001 & 0.001 & 0.000 & 0.000 & 0.001 \\
MT & 0.000 & 0.000 & 0.000 & 0.000 & 0.000 & 0.000 & 0.000 & 0.000 \\
LDM & 0.901 & 0.930 & 0.874 & 0.936 & 0.834 & 0.849 & 0.831 & 0.884 \\
DDM & 0.003 & 0.002 & 0.002 & 0.003 & 0.004 & 0.007 & 0.003 & 0.007 \\
DEF & 0.003 & 0.002 & 0.002 & 0.001 & 0.001 & 0.002 & 0.001 & 0.211 \\
LDM-WB & 0.235 & 0.285 & 0.258 & 0.252 & 0.140 & 0.214 & 0.120 & 0.341 \\
\midrule
Spearman $\rho$ vs. default & 0.99 & 1.00 & 1.00 & 0.99 & 0.99 & 0.97 & 0.93 & 0.96 \\
\bottomrule
\end{tabular}
}
\caption{\textbf{Gender}: cDFS under eight discriminator configurations, horizon $L=32$. Spearman $\rho$ compares the model ranking under each configuration with the ranking under GRU-512-2L.}
\label{tab:detector-gender}
\end{table}

\begin{table}[ht]
\centering
\resizebox{\textwidth}{!}{%
\begin{tabular}{lcccccccc}
\toprule
Method & \makecell{GRU\\128-1L} & \makecell{GRU\\512-1L} & \makecell{GRU\\512-2L (default)} & \makecell{GRU\\1024-3L} & \makecell{Tr.\\128-1L} & \makecell{Tr.\\512-1L} & \makecell{Tr.\\512-2L} & \makecell{Tr.\\1024-3L} \\
\midrule
GT & 1.000 & 1.000 & 1.000 & 1.000 & 1.000 & 1.000 & 1.000 & 1.000 \\
Hist. & 0.573 & 0.556 & 0.544 & 0.550 & 0.338 & 0.313 & 0.147 & 0.874 \\
Mode & 0.000 & 0.000 & 0.000 & 0.000 & 0.000 & 0.000 & 0.000 & 0.000 \\
Rep. & 0.015 & 0.011 & 0.009 & 0.006 & 0.000 & 0.002 & 0.000 & 0.494 \\
AR & 0.000 & 0.000 & 0.000 & 0.000 & 0.000 & 0.000 & 0.000 & 0.000 \\
MT & 0.000 & 0.000 & 0.000 & 0.000 & 0.000 & 0.000 & 0.000 & 0.000 \\
LDM & 0.867 & 0.866 & 0.855 & 0.832 & 0.810 & 0.836 & 0.811 & 0.929 \\
DDM & 0.004 & 0.004 & 0.004 & 0.005 & 0.005 & 0.008 & 0.005 & 0.021 \\
DEF & 0.000 & 0.000 & 0.000 & 0.000 & 0.000 & 0.000 & 0.000 & 0.203 \\
LDM-WB & 0.161 & 0.160 & 0.175 & 0.178 & 0.111 & 0.178 & 0.050 & 0.606 \\
\midrule
Spearman $\rho$ vs. default & 1.00 & 1.00 & 1.00 & 1.00 & 0.93 & 0.99 & 0.93 & 0.95 \\
\bottomrule
\end{tabular}
}
\caption{\textbf{Alphabattle}: cDFS under eight discriminator configurations, horizon $L=32$. Spearman $\rho$ compares the model ranking under each configuration with the ranking under GRU-512-2L.}
\label{tab:detector-alphabattle}
\end{table}

\subsection{Shape and Trend scores}\label{app:realism:shape-trend}

As auxiliary distributional diagnostics, we report Shape and Trend from~\citet{tabsyn}, computed via the \texttt{sdmetrics} \texttt{QualityReport} on the pool of predicted and real continuations in the evaluation set.
Shape aggregates Kolmogorov--Smirnov distances (for numerical features) and total-variation distances (for categorical features) between the marginal distributions of each generated and real feature.
Trend aggregates disagreements between pairwise feature dependencies (correlations for numerical pairs, contingency-table distances for categorical pairs).
Both scores are bounded in $[0, 1]$, with $1$ indicating perfect agreement.
In contrast to cDFS, Shape and Trend ignore event ordering and history, so they quantify only marginal and low-order joint fidelity of features pooled across sequences.
We include them as a sanity check on per-feature distribution quality rather than as a primary realism measure.
Table~\ref{tab:shape-trend-h32} reports Shape and Trend alongside cDFS for all models at horizon $L=32$.
Within the cluster that cDFS maps to near zero, the discriminator-free Shape and Trend scores still separate models by a large margin (e.g., on Age: 0.79/0.40 for Mode vs. 0.90/0.61 for DEF vs. 0.99/0.97 for Repeat), providing a graded signal where the detector saturates.
Conversely, Repeat shows that near-perfect Shape and Trend do not imply realism: the detector identifies its continuations almost solely through the verbatim repetition of past events.
The relationship between cDFS and the two scores is analyzed in Appendix~\ref{app:more-res:metric-corr}.

\begin{table}[ht]
\centering
\resizebox{\textwidth}{!}{%
\begin{tabular}{lccccccccc}
\toprule
 & \multicolumn{3}{c}{cDFS $\uparrow$} & \multicolumn{3}{c}{Shape $\uparrow$} & \multicolumn{3}{c}{Trend $\uparrow$} \\
\cmidrule(lr){2-4}\cmidrule(lr){5-7}\cmidrule(lr){8-10}
Method & Age & Gender & Alph. & Age & Gender & Alph. & Age & Gender & Alph. \\
\midrule
Hist. & \textbf{0.234} & \textbf{0.476} & \textbf{0.544} & \textbf{0.987} & \textbf{0.962} & \textbf{0.984} & \textbf{0.961} & \textbf{0.785} & \textbf{0.949} \\
Mode & 0.000 & 0.000 & 0.000 & 0.792 & 0.806 & 0.801 & 0.402 & 0.700 & 0.567 \\
Rep. & 0.065 & 0.050 & 0.009 & 0.994 & 0.978 & 0.991 & 0.973 & 0.879 & 0.955 \\
AR & 0.000 & 0.002 & 0.000 & 0.852 & 0.851 & 0.840 & 0.538 & 0.642 & 0.620 \\
MT & 0.000 & 0.000 & 0.000 & 0.853 & 0.811 & 0.855 & 0.401 & 0.272 & 0.712 \\
LDM & \textbf{1.000} & \textbf{0.907} & \textbf{0.856} & \textbf{0.989} & \textbf{0.960} & \textbf{0.977} & \textbf{0.950} & \textbf{0.926} & \textbf{0.914} \\
DDM & \textbf{0.202} & 0.002 & 0.004 & \textbf{0.966} & 0.695 & 0.740 & \textbf{0.888} & 0.567 & 0.603 \\
DEF & 0.000 & 0.002 & 0.000 & 0.902 & 0.876 & 0.884 & 0.607 & 0.704 & 0.563 \\
LDM-WB & \textbf{0.224} & \textbf{0.256} & \textbf{0.177} & \textbf{0.952} & \textbf{0.938} & \textbf{0.943} & \textbf{0.801} & \textbf{0.891} & \textbf{0.868} \\
DDM-WB & \textbf{0.102} & 0.001 & 0.001 & \textbf{0.941} & 0.709 & 0.733 & \textbf{0.711} & 0.613 & 0.579 \\
\bottomrule
\end{tabular}
}
\caption{Shape and Trend scores alongside cDFS at horizon $L=32$ on the three transaction datasets. Bold marks cDFS $> 0.07$ and the corresponding Shape and Trend cells.}
\label{tab:shape-trend-h32}
\end{table}

\section{Additional Experiments}\label{app:more-res}

\subsection{Replacing cDFS with simpler diagnostics.}\label{app:more-res:metric-corr}

\begin{figure}[ht]
    \centering
    \includegraphics[width=0.9\linewidth]{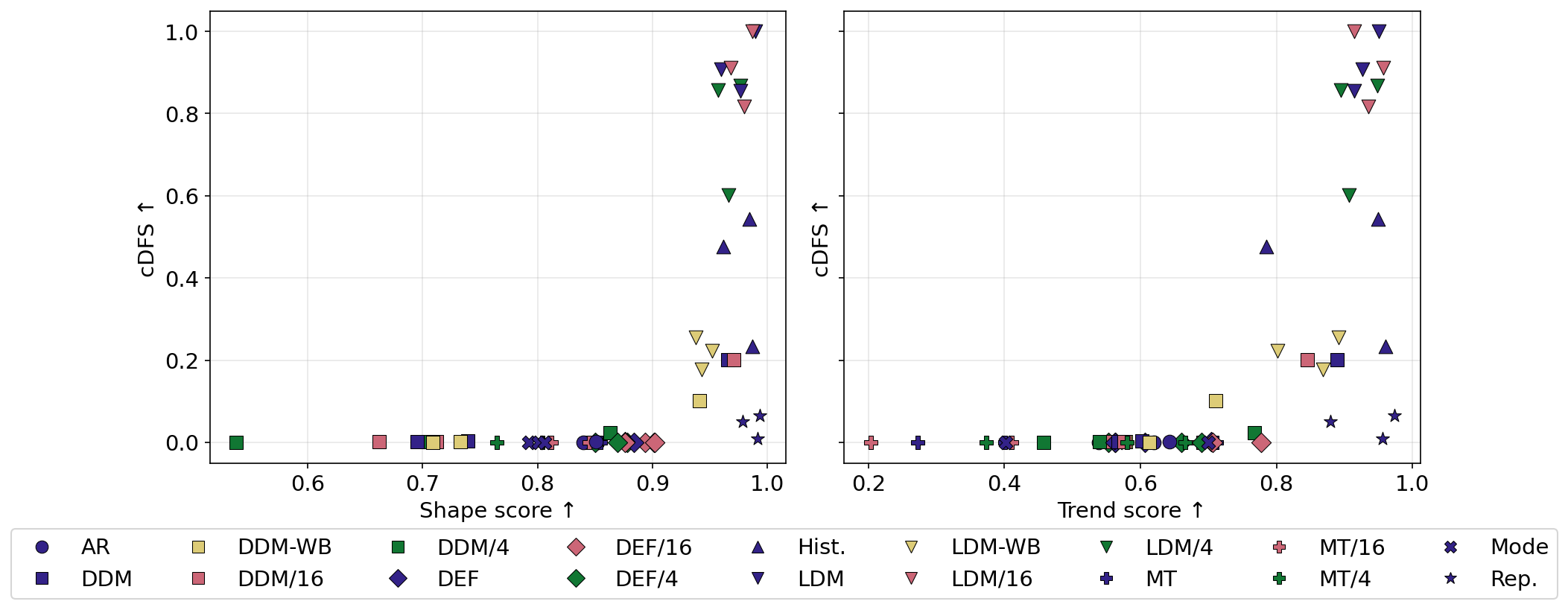}
    \caption{%
    Relationship between cDFS and the auxiliary fidelity scores — Shape score (left) and Trend score (right) — across all three datasets pooled.
    Horizon $L=32$.
    Higher is better on both axes.
    The figure shows that high Shape and Trend scores do not, by themselves, imply a high cDFS: many baselines achieve competitive Shape or Trend values while collapsing along the cDFS axis, suggesting that cDFS is a more holistic criterion.
    }
    \label{fig:shape-trend}
\end{figure}

cDFS is computationally heavier than per-feature distributional scores --- it requires training and evaluating a discriminator --- so it is natural to ask whether the cheaper Shape and Trend statistics from~\citet{tabsyn} suffice as a substitute.
Figure~\ref{fig:shape-trend} pools predictions from all three datasets at $L=32$ and plots cDFS against each of the two scores.
The relationship is strikingly asymmetric: a high Shape or Trend score implies nothing about cDFS, but a high cDFS does require high Shape and Trend.
In other words, Shape and Trend agreement is necessary but far from sufficient for distributional realism.
The reason is structural.
Shape and Trend pool features across all sequence positions, ignoring event order, conditioning on history, and joint dependencies of more than two features.
A model that produces marginally and pairwise plausible rows but mis-orders them in time, or generates implausible continuations of a given history, can score well on Shape and Trend while being trivially detected by cDFS.
Shape and Trend are useful auxiliary diagnostics, but cDFS captures sequential and conditional structure that they miss, and is therefore the appropriate primary realism criterion for event-sequence forecasting.

\subsection{Wasserstein BaryBooster Independent Evaluation}\label{app:more-res:wb-ind-eval}

In the main text (\S\ref{sec:results}), we show that BaryBoosted diffusion models achieve SoTA TMS values.
We conclude that diffusion models can produce accurate predictions with the proposed technique.
However, the algorithm is designed to optimize the same metric used for its evaluation, potentially leading to biased results.
In this section, we support the broad claim of high prediction accuracy by reporting independent metrics that BaryBooster does not optimize directly.

Table~\ref{tab:more_metrics_h32} shows that the application of Wasserstein BaryBooster on top of LDM improves all prediction accuracy metrics consistently across datasets.
Despite its poor realism, DDM also shows improved accuracy after BaryBooster application, with a few exceptions.
These observations support our broad claim of improved prediction accuracy.

\begin{table}[ht]
\centering
\caption{TMS, R1 (amount), R1 (time), F1 (mcc) at horizon $L=32$. Across all shown methods, the best is in \textbf{bold}, the second best is \uline{underlined} and the third best is \dashuline{dash-underlined}.}
\label{tab:more_metrics_h32}
\resizebox{\textwidth}{!}{%
\begin{tabular}{lcccccccccccc}
\toprule
Method & \multicolumn{3}{c}{TMS ↑} & \multicolumn{3}{c}{R1 (amount) ↑} & \multicolumn{3}{c}{R1 (time) ↑} & \multicolumn{3}{c}{F1 (mcc) ↑} \\
\cmidrule(lr){2-4} \cmidrule(lr){5-7} \cmidrule(lr){8-10} \cmidrule(lr){11-13}
 & Alphabattle & Gender & Age & Alphabattle & Gender & Age & Alphabattle & Gender & Age & Alphabattle & Gender & Age \\
\midrule
Hist. & 0.382 & 0.221 & 0.366 & \dashuline{0.498} & 0.117 & 0.245 & -0.049 & -0.325 & -0.179 & 0.424 & 0.397 & 0.331 \\
Mode & 0.203 & 0.047 & 0.263 & 0.000 & 0.000 & 0.000 & 0.000 & 0.000 & 0.000 & 0.081 & 0.088 & 0.052 \\
Rep. & \underline{0.440} & \dashuline{0.361} & 0.454 & \underline{0.524} & \dashuline{0.182} & 0.236 & 0.147 & 0.137 & 0.128 & \underline{0.444} & \underline{0.425} & \textbf{0.374} \\
AR & 0.286 & 0.188 & 0.354 & 0.043 & -0.018 & 0.082 & 0.064 & 0.096 & 0.183 & 0.287 & 0.324 & 0.208 \\
MT & 0.286 & 0.153 & 0.360 & -0.000 & 0.016 & 0.020 & -0.042 & -0.043 & 0.185 & 0.369 & 0.343 & 0.270 \\
LDM & 0.405 & 0.318 & 0.426 & 0.479 & 0.180 & 0.216 & \dashuline{0.148} & \dashuline{0.144} & 0.074 & 0.400 & 0.364 & 0.323 \\
DDM & 0.251 & 0.107 & 0.441 & -0.015 & 0.059 & 0.301 & 0.125 & 0.071 & 0.203 & 0.275 & 0.238 & 0.299 \\
DEF & \dashuline{0.424} & \underline{0.413} & \underline{0.508} & 0.364 & \underline{0.287} & \dashuline{0.331} & \underline{0.294} & \underline{0.324} & \textbf{0.353} & \dashuline{0.440} & \dashuline{0.413} & \dashuline{0.360} \\
\midrule
LDM-WB (ours) & \textbf{0.495} & \textbf{0.436} & \textbf{0.522} & \textbf{0.583} & \textbf{0.334} & \textbf{0.438} & \textbf{0.337} & \textbf{0.331} & \underline{0.331} & \textbf{0.459} & \textbf{0.448} & \underline{0.371} \\
DDM-WB (ours) & 0.274 & 0.136 & \dashuline{0.480} & 0.086 & 0.070 & \underline{0.373} & 0.006 & 0.047 & \dashuline{0.240} & 0.285 & 0.264 & 0.358 \\
\bottomrule
\end{tabular}
}
\end{table}

\subsection{TMS and matched submetrics}\label{app:more-res:tms-submetrics}

\begin{figure}[ht]
    \centering
    \includegraphics[width=\linewidth]{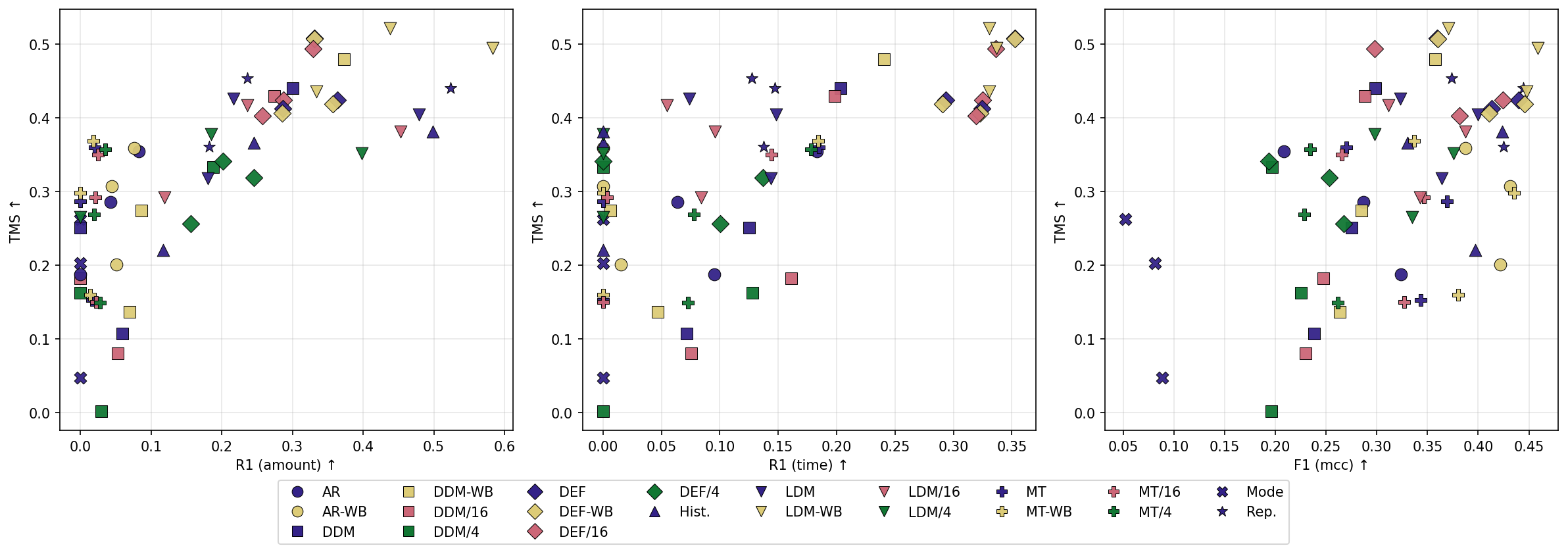}
    \caption{%
    Relationship between TMS and matched per-feature submetrics at horizon $L=32$: matched $R_1$ for amount (left), matched $R_1$ for time (middle), and matched $F_1$ for MCC (right).
    Higher is better on both axes.
    Across all three views, TMS grows consistently with each matched submetric.
    }
    \label{fig:tms-submetrics}
\end{figure}

Figure~\ref{fig:tms-submetrics} shows a clear positive correlation between TMS and matched $R_1$ for amount, matched $R_1$ for time, and matched $F_1$ for MCC.
For clarity, these matched submetrics are \emph{separately computed} feature-wise metrics ($R_1^{m,\star}$, $R_1^{t,\star}$, $F_1^{n,\star}$; Appendix~\ref{app:tms:matched}), each using its own best permutation, rather than direct components of TMS.
TMS instead uses a \emph{single shared} permutation across all features (Appendix~\ref{app:tms:tms}); the observed correlation therefore indicates that TMS aligns well with common, easy-to-interpret per-feature metrics and that TMS gains reflect real quality improvements across forecasting dimensions.

\subsection{Other Horizons}\label{app:more-res:horizon}

\begin{figure}
    \centering
    \includegraphics[width=\linewidth]{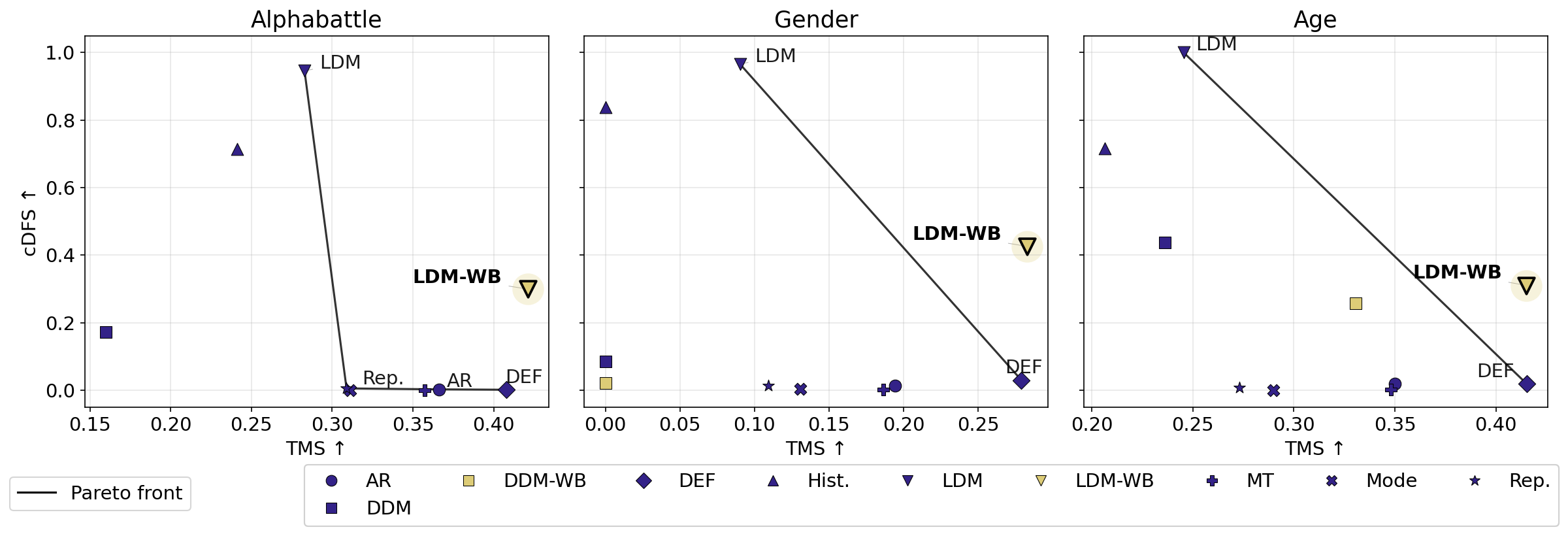}
    \caption{%
        Trade-off between TMS and cDFS for each dataset on horizon $L=4$.
        The solid line shows the Pareto front over baseline methods.
        }
    \label{fig:pareto_4}
\end{figure}

\begin{figure}
    \centering
    \includegraphics[width=\linewidth]{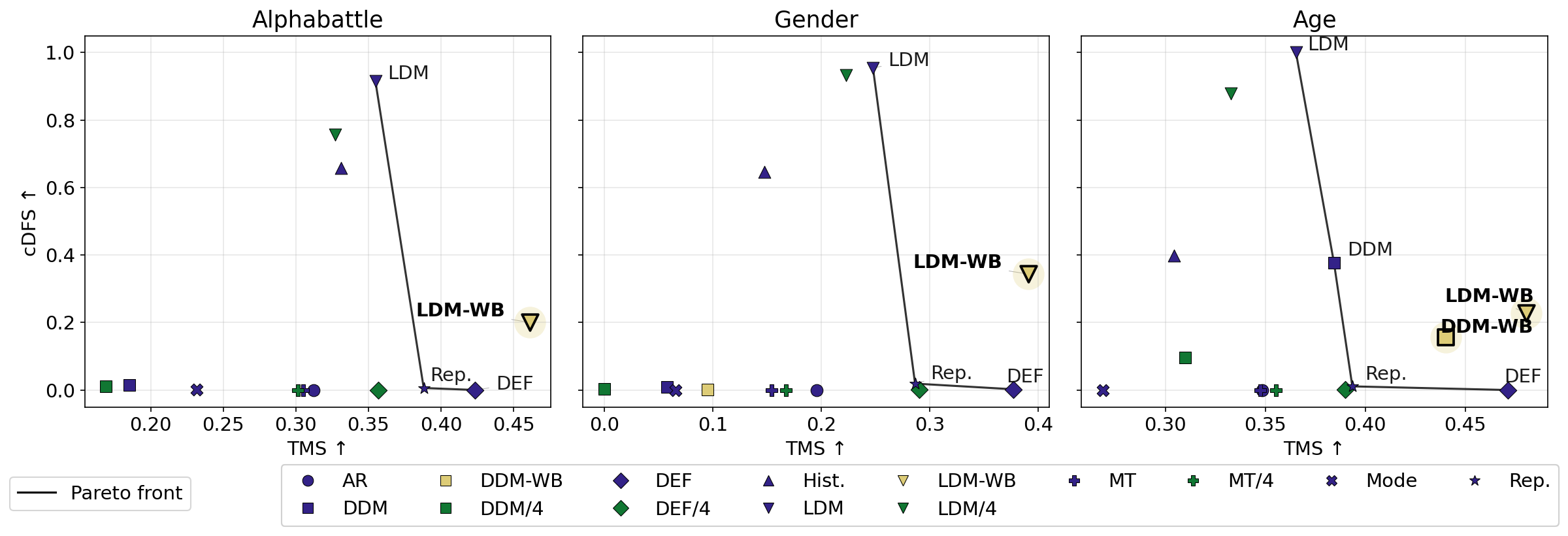}
    \caption{%
        Trade-off between TMS and cDFS for each dataset on horizon $L=16$.
        The solid line shows the Pareto front over baseline methods.
        }
    \label{fig:pareto_16}
\end{figure}

\begin{figure}
    \centering
    \includegraphics[width=\linewidth]{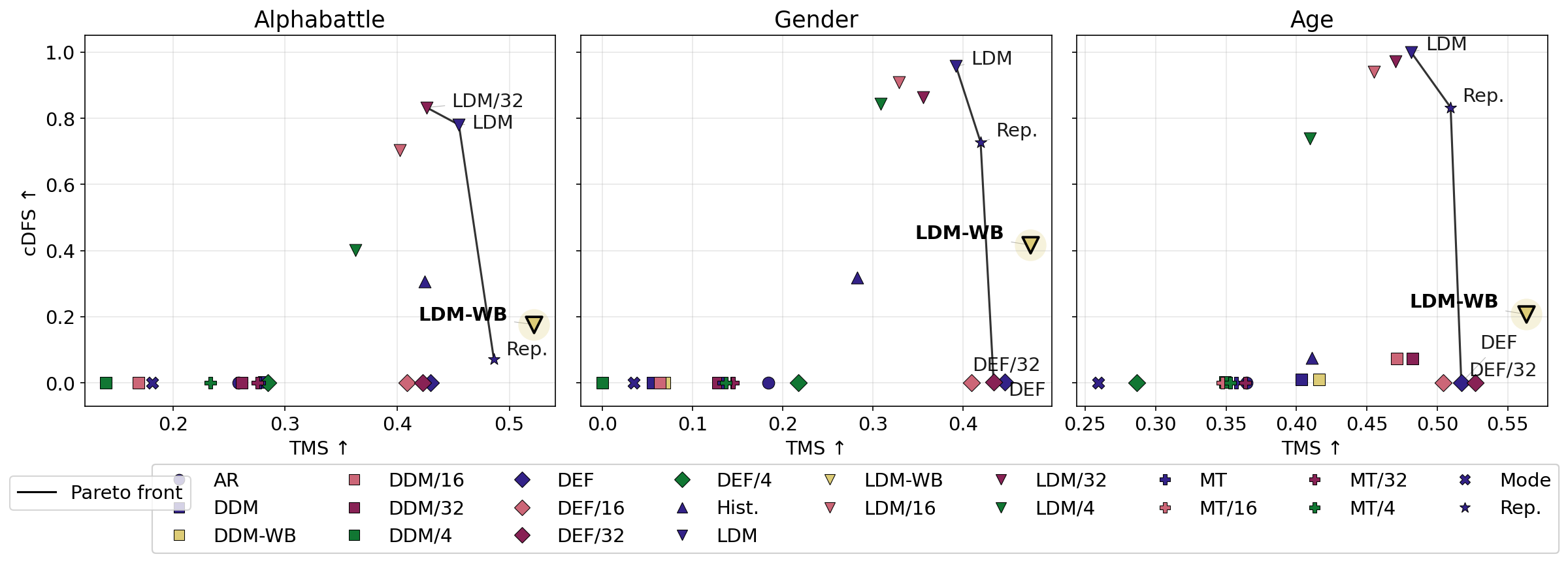}
    \caption{%
        Trade-off between TMS and cDFS for each dataset on horizon $L=64$.
        The solid line shows the Pareto front over baseline methods.
        }
    \label{fig:pareto_64}
\end{figure}

In this section, we present experimental results with horizons $L \in \{4, 16, 64\}$ (Figures~\ref{fig:pareto_4}, \ref{fig:pareto_16}, and~\ref{fig:pareto_64}).

At $L=64$, Repeat receives a high cDFS on Gender and Age (0.73 and 0.83). As in Appendix~\ref{app:realism:detector-ablation}, we attribute this to the detector not catching the verbatim copy of the history at this horizon, not to realistic continuations, so the cDFS of Repeat at $L=64$ should be read with caution. The rest of the picture matches $L=32$: LDM is the most realistic model, and LDM-WB has the highest TMS on all datasets.

\subsection{Statistical Significance}\label{app:more-res:seeds}

\begin{table}[ht]
\centering
\caption{Age, horizon $L=32$: 3-seed mean $\pm$ std vs. reported in main results (single reported values shown without uncertainty). Higher is better. Within each column, the best is in \textbf{bold}, second best is \uline{underlined}}
\label{tab:metrics_h32_age_seeds_reported}
\begin{tabular}{lcccc}
\toprule
Method & \multicolumn{2}{c}{TMS ↑} & \multicolumn{2}{c}{cDFS ↑} \\
 & 3-seed & reported & 3-seed & reported \\
\midrule
LDM & {0.430 $\pm$ 0.002} & {0.426} & \textbf{0.983 $\pm$ 0.006} & \textbf{1.000} \\
DEF & \uline{0.505 $\pm$ 0.004} & \uline{0.508} & {0.001 $\pm$ 0.000} & {0.000} \\
\midrule
LDM-WB & \textbf{0.523 $\pm$ 0.001} & \textbf{0.522} & \uline{0.202 $\pm$ 0.005} & \uline{0.224} \\
\bottomrule
\end{tabular}
\end{table}

In this experiment, we show that the metrics are stable across runs and justify our choice not to run each experiment multiple times.
We report mean and standard deviations over three random seeds in Table~\ref{tab:metrics_h32_age_seeds_reported} for the spotlight methods.
We draw our conclusions on metric gaps that are much larger than standard deviations across runs, so we do not question the statistical significance.
One entry of Table~\ref{tab:metrics_h32_age_seeds_reported} deserves a
comment: the reported cDFS of LDM-WB (0.224) lies several across-seed standard
deviations away from the 3-seed mean ($0.202 \pm 0.005$). The across-seed
deviation, however, is only one component of the variance: each cDFS value is
itself the mean of a 3-fold cross-validation of the detector, with an internal
standard deviation of ${\approx}0.02$ (0.024 for LDM-WB), so the gap is within
one standard deviation of the full estimate. The detector is inherently noisier
in the intermediate score range, while the discriminator-free Shape and Trend
scores are stable there.

\subsection{Qualitative Samples}\label{app:more-res:qualitative}

\begin{figure}
    \centering
    \includegraphics[width=\linewidth]{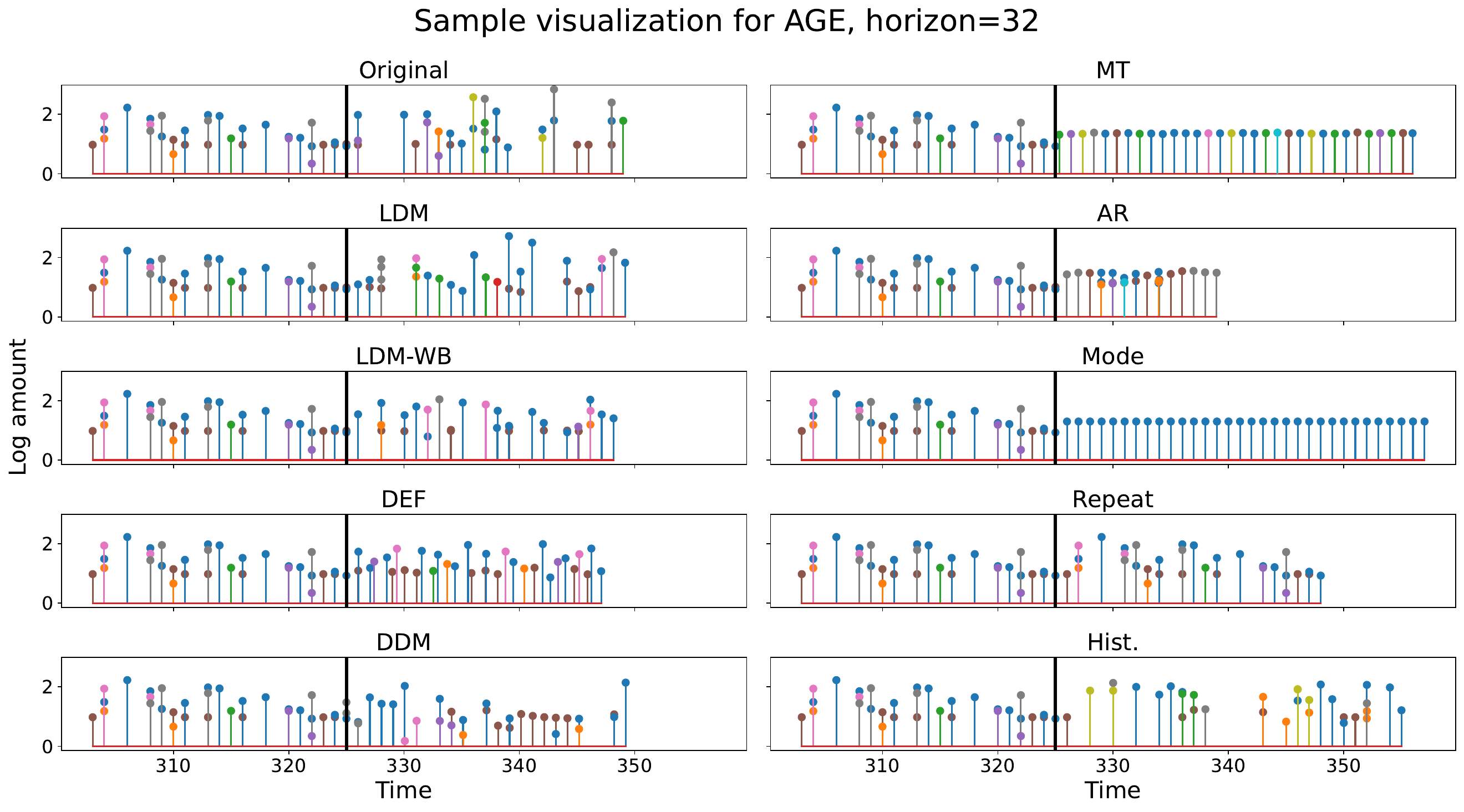}
    \caption{%
        Generated continuations for one Age test client at horizon $L=32$.
        Each event is a stem at its timestamp; height is the log-transformed amount and
        color encodes the category (the nine most frequent categories receive their own
        color, the rest share grey). The vertical rule separates the observed history,
        identical across all panels, from the 32 generated events. \emph{Original} is
        the ground-truth continuation.
        }
    \label{fig:samples_age_32}
\end{figure}

Figure~\ref{fig:samples_age_32} shows generated continuations for one Age test
client at horizon $L=32$. All methods are conditioned on the identical history, so the
panels differ only in what each model generates.

Several failure modes are directly visible. MT and AR keep a plausible
mix of categories, but their numerical features collapse: the amounts sit at a
nearly constant value and the inter-arrival times form an almost uniform grid
--- the conditional-mean pathology of Figure~\ref{fig:intro-example}. Mode
degenerates into a single repeated event, and Repeat copies the history
verbatim. LDM is the only model whose continuation reproduces both the
irregular timing and the amount variability of the real one; HistSampler
matches the marginals but places events at implausible times.

\subsection{Comparison of LDM with ADiff4TPP}\label{app:more-res:adiff4tpp}

\begin{table}[ht]
\centering
\caption{ADiff4TPP and LDM at horizon $L=32$ on Age dataset, with and without Wasserstein BaryBooster.}
\label{tab:metrics_age_adiff_ldm_h32}
\begin{tabular}{lcc}
\toprule
Method & TMS ↑ & cDFS ↑ \\
\midrule
ADiff4TPP & {0.402} & {0.997} \\
LDM & \textbf{0.426} & \textbf{1.000} \\
\midrule
ADiff4TPP-WB & {0.512} & 0.108 \\
LDM-WB & \textbf{0.522} & \textbf{0.224} \\
\bottomrule
\end{tabular}
\end{table}

Table~\ref{tab:metrics_age_adiff_ldm_h32} shows that our implementation of the latent diffusion model performs on par with or slightly better than ADiff4TPP~\citep{adiff4tpp}, so we do not include it as a separate model in the comparison.

\section{Wasserstein BaryBooster: optimization details}\label{app:barybooster}

We solve equation~\eqref{eq:barybooster} by alternating minimization (Algorithm~\ref{alg:barybooster}), similar to consensus decoding for event sequences~\citep{mei2019imputing}.
With $\hat S$ fixed, the optimal alignment to each $\hat S_n$ is recovered exactly by Hungarian matching~(Appendix~\ref{app:tms:hungarian}).
With the alignments fixed, each feature of $\hat S$ admits a closed-form coordinate-wise update: time and numerical features are set to the median of the aligned values (the Fr\'echet mean under the absolute-deviation cost of \ref{app:tms:r1}); categorical features are set to the mode (the Fr\'echet mean under the $0/1$ cost of \ref{app:tms:f1}).
We initialize $\hat S$ with the medoid of $\{\hat S_n\}_{n=1}^N$, the sample minimizing the sum of pairwise TMS costs to the others.
For categorical features, the matching uses the $0/1$ cost as a surrogate of the F1 term, so the procedure optimizes this surrogate objective and finds a local solution.
Each iteration weakly decreases the objective, and the alignment set is finite, so the procedure terminates in finitely many steps.

\begin{algorithm}[t]
\caption{Wasserstein BaryBooster}\label{alg:barybooster}
\begin{algorithmic}[1]
\Require base generator $G$, history $S$, sample count $N$, max iterations $T$
\State sample $\hat S_n \sim G(\cdot \mid S)$ for $n = 1, \ldots, N$
\State $n^\star \gets \arg\min_{n \in [N]} \sum_{m=1}^{N} d_{\mathrm{TMS}}\!\left(\hat S_n, \hat S_m\right)$;
       $\hat S \gets \hat S_{n^\star}$ \Comment{medoid init}
\For{$t = 1, \ldots, T$}
    \For{$n = 1, \ldots, N$}
        \State $\sigma_n \gets \arg\min_{\sigma} \sum_{i \in \mathcal{H}} \bar C_{(\hat S,\, \hat S_n)}(i, \sigma(i))$
               \Comment{Hungarian}
    \EndFor
    \If{$t>1$ \textbf{and} $\sigma_n = \sigma_n^{\mathrm{prev}}$ for all $n \in [N]$}
        \State \textbf{break}
    \EndIf
    \State $\sigma_n^{\mathrm{prev}} \gets \sigma_n$ for all $n \in [N]$
    \For{$i \in \mathcal{H}$}
        \State $\hat t_i \gets \mathrm{med}\bigl\{ t^{(n)}_{\sigma_n(i)} : n=1,\ldots,N \bigr\}$
        \For{$m = 1, \ldots, M$}
            \State $\hat x_i^{m} \gets \mathrm{med}\bigl\{ x^{m,(n)}_{\sigma_n(i)} : n=1,\ldots,N \bigr\}$
        \EndFor
        \For{$q = 1, \ldots, N_{\mathrm{cat}}$}
            \State $\hat k_i^{q} \gets \operatorname{mode}\bigl\{ k^{q,(n)}_{\sigma_n(i)} : n=1,\ldots,N \bigr\}$
        \EndFor
    \EndFor
\EndFor
\State \textbf{return} $\hat S$
\end{algorithmic}
\end{algorithm}

\subsection{Effect of the sample size \texorpdfstring{$N$}{N}}\label{app:barybooster-n}

Table~\ref{tab:barybooster-n} reports LDM-WB at horizon $L=32$ as a
function of the number of aggregated samples $N$; $N=1$ corresponds to the base
LDM without boosting. Already $N=5$ recovers about $75\%$ of the full TMS gain
at one sixth of the sampling cost, $N=10$ recovers about $90\%$, and the curve
saturates after $N \approx 10$--$15$.

\begin{table}[ht]
\centering
\begin{tabular}{lcccccc}
\toprule
 & \multicolumn{2}{c}{Age} & \multicolumn{2}{c}{Gender} & \multicolumn{2}{c}{Alphabattle} \\
\cmidrule(lr){2-3}\cmidrule(lr){4-5}\cmidrule(lr){6-7}
$N$ & TMS $\uparrow$ & cDFS $\uparrow$ & TMS $\uparrow$ & cDFS $\uparrow$ & TMS $\uparrow$ & cDFS $\uparrow$ \\
\midrule
1  & 0.428 & 0.966 & 0.319 & 0.874 & 0.404 & 0.855 \\
5  & 0.499 & 0.257 & 0.407 & 0.137 & 0.472 & 0.133 \\
10 & 0.510 & 0.271 & 0.425 & 0.300 & 0.485 & 0.210 \\
15 & 0.517 & 0.229 & 0.430 & 0.271 & 0.491 & 0.188 \\
30 & 0.522 & 0.224 & 0.436 & 0.258 & 0.495 & 0.177 \\
\bottomrule
\end{tabular}
\caption{Wasserstein BaryBooster on LDM as a function of the sample size $N$ at horizon $L=32$; $N{=}1$ is the base LDM without boosting.}
\label{tab:barybooster-n}
\end{table}

\section{Wasserstein BaryBooster on remaining models}\label{app:barybooster-other}

\begin{figure}[ht]
    \centering
    \includegraphics[width=0.9\linewidth]{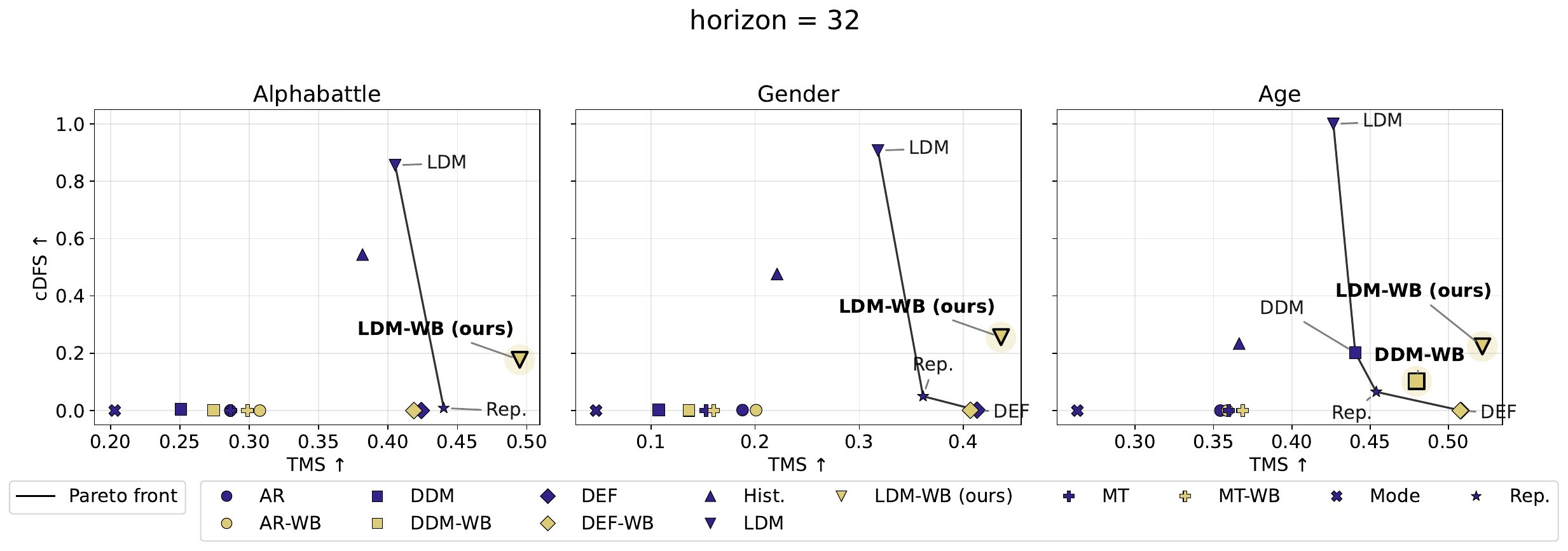}
    \caption{%
    TMS change at horizon $L=32$ after applying Wasserstein BaryBooster to different base models.
    The largest improvement is observed for diffusion-based models, while DEF shows no visible gain and the remaining methods improve only slightly.
    }
    \label{fig:boosters-32}
\end{figure}

Figure~\ref{fig:boosters-32} confirms that BaryBooster is most effective when the base generator is diffusion-based: the boost is large for diffusion models, absent for DEF, and small for the other non-diffusion methods.

\section{Preprocessing Details}
\label{app:preprocessing}

Across all datasets, we apply \texttt{log1p} to the transaction amount.
We also split the data into three subsets: ``train'' for neural network training, ``val'' for early stopping, and ``test'' for reporting the metrics.
The test split is out-of-sample, and the validation split is randomly sampled from the training data.
During training, we apply a diff transform to the time feature, but we restore it back to absolute time for metrics calculation.

\paragraph{Split strategy.}
\textbf{Age} uses a stratified user split by bins with a $20\%$ test fraction.
\textbf{Gender} uses a random user split with a $20\%$ test fraction.
\textbf{Alphabattle} uses separate raw train/test populations, with the test population strictly later than the train population.

\paragraph{Dataset-specific filtering.}
\textbf{Gender} keeps only sequences with length in $[20, 1500)$, maximum \texttt{hour\_diff} at most $100$, and median \texttt{hour\_diff} at most $10$.
\textbf{Alphabattle} keeps only sequences with length in $[128, 1200]$, maximum \texttt{hour\_diff} below $800$, median \texttt{hour\_diff} below $70$, drops clients with rare \texttt{payment\_system}, \texttt{mcc\_category}, and \texttt{card\_type} values, restricts to \texttt{currency = 1}, and is subsampled to 100K users for computational efficiency.

\paragraph{History--continuation cutoffs.}
During training and validation we sample a cutoff time uniformly at random over an interval of admissible times for each sequence; events strictly before the cutoff form the history, and the next $L$ events form the target continuation.
The same cutoff protocol applies to training, validation, and evaluation; during training a new cutoff is sampled every epoch, whereas validation and test use fixed cutoffs per sequence for the entire study.
For validation and test we use one fixed cutoff per sequence on \textbf{Age} and \textbf{Alphabattle}.
On \textbf{Gender} (our smallest dataset) we instead use three fixed cutoffs per sequence, spreading evaluation contexts to reduce metric variance.
After sampling cutoffs, we keep only sequences with at least $L$ events on the history side.
Using random cutoffs during training lets the model forecast from diverse time periods rather than repeatedly from the same calendar segment.

\section{Implementation Details}
\label{app:methods}

This appendix records implementation and training details omitted from \S\ref{sec:methods}, including optimization and validation protocols, shared backbones, diffusion schedules, and non-learned baselines.

\subsection{Comparison of LDM and ADiff4TPP~\citep{adiff4tpp}}\label{app:methods:ldm-vs-adiff4tpp}

\begin{table}[ht]
\centering
\caption{High-level comparison between ADiff4TPP and our latent diffusion model (LDM).}
\label{tab:ldm-vs-adiff4tpp}
\begin{tabular}{lll}
\toprule
Aspect & ADiff4TPP~\citep{adiff4tpp} & Our LDM \\
\midrule
Diffusion formulation & Flow matching & EDM \\
Backbone & DiT & DiT \\
Data--latent mapping & $\beta$-VAE & $\beta$-VAE \\
\bottomrule
\end{tabular}
\end{table}

In the original ADiff4TPP setup, forecasting is done with an \emph{unconditional} model. For an apples-to-apples empirical comparison, we train \emph{conditional} variants for both our LDM and our re-implementation of ADiff4TPP. Additionally, we apply classifier-free guidance (CFG) to our LDM at inference time (\S\ref{app:shared}).

\subsection{Schematic Illustration of Methods}

\begin{figure}[ht]
    \centering

    % Top subfigure
    \begin{subfigure}{\linewidth}
        \centering
        \includegraphics[width=0.99\linewidth]{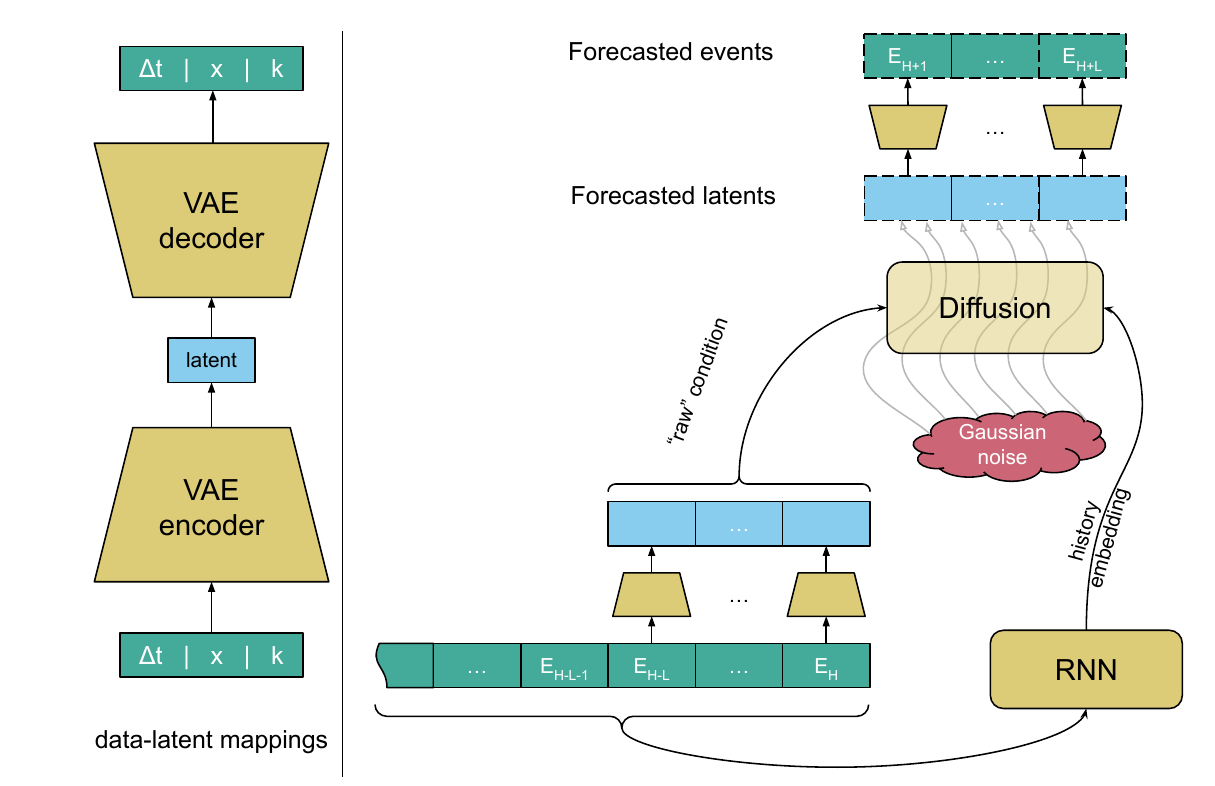}
        \caption{Latent Diffusion Model (LDM)}
        \label{fig:ldm}
    \end{subfigure}

    \vspace{0.5em}

    % Bottom two subfigures side by side
    \begin{subfigure}{0.49\linewidth}
        \centering
        \includegraphics[width=\linewidth]{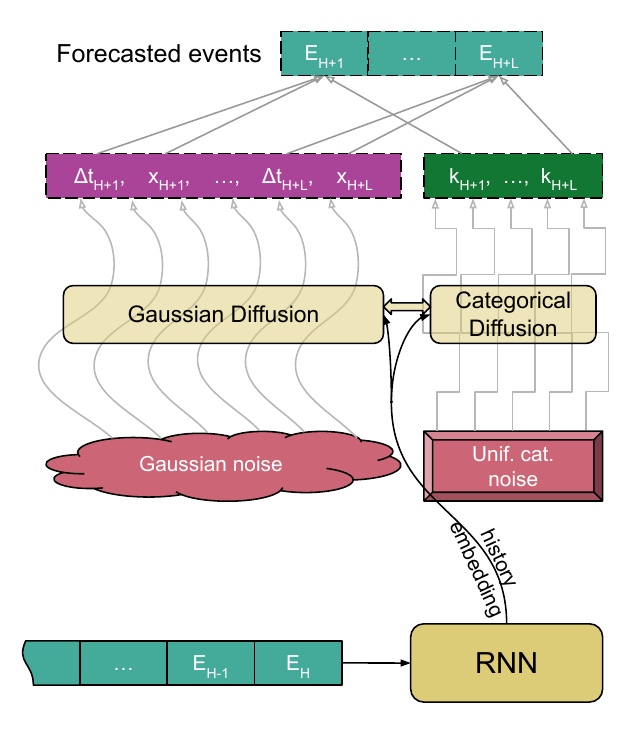}
        \caption{Data Diffusion Model (DDM)}
        \label{fig:ddm}
    \end{subfigure}
    \hfill
    \begin{subfigure}{0.49\linewidth}
        \centering
        \includegraphics[width=\linewidth]{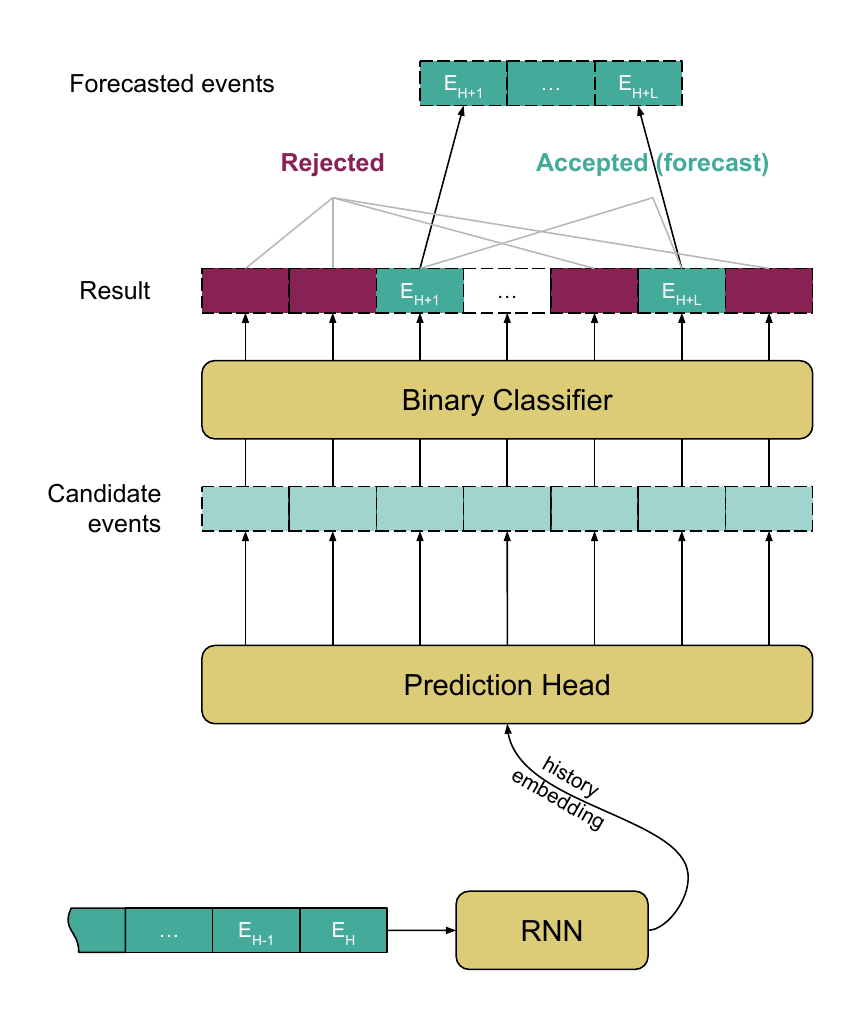}
        \caption{DEF}
        \label{fig:def}
    \end{subfigure}

    \caption{Schematic illustration of the considered methods.}
    \label{fig:methods}
\end{figure}

Figure~\ref{fig:methods} illustrates the three adapted methods: LDM, DDM, and DEF.

\subsection{Shared protocol and infrastructure}
\label{app:shared}

\paragraph{Classifier-free guidance (CFG).} CFG is a well-established and popular tool \citep{cfg} which allows adjusting (conditional) diffusion models on \textit{diversity}-\textit{fidelity} coordinate plane, enabling practitioners to balance sample diversity against generation fidelity by modulating the strength ($w \geq 0$) of the conditioning signal during inference. Technically speaking, CFG prescribes simultaneous training of conditional and unconditional denoising models (parameterized as a single model, unconditional training appears by substituting the actual class with the ``empty'' class $\emptyset$  with probability $p_{\text{uncond}}$) and subsequent \textit{guidance} of the usual conditional model with extra conditioning at inference stage: 
\begin{gather*}
    \text{Denoiser}_{\text{CFG}}(\text{data}, \text{condition}) = (1 + w)\text{Denoiser}(\text{data}, \text{condition}) - w \text{Denoiser}(\text{data}, \emptyset).
\end{gather*}
This technique was found to be promising for event sequence generation \citep{tabdit}, and we adopt it for our LDM and DDM models. 

\paragraph{Framework and reproducibility.}
All neural models are implemented in PyTorch.
Unless stated otherwise below, training and evaluation follow a single shared protocol.
Randomness is controlled with a single global seed.

\paragraph{Optimization.}
We optimize with AdamW (weight decay $0.01$), clip gradients at global norm $1$, and search the learning rate on a logarithmic grid from $10^{-5}$ to $3 \times 10^{-3}$.
Training may run for up to one million iterations with early stopping when validation loss plateaus (patience $20$ epochs, or $10$ for \textbf{Alphabattle}).
For latent diffusion only, we apply exponential smoothing to the validation loss before comparing it to the running optimum.

\paragraph{Weight tracking and batching.}
We maintain an exponential moving average of model parameters with decay $0.9999$, applied every ten steps after $70{,}000$ training steps.
Unless noted, the batch size is $128$; \textbf{DEF} and \textbf{AR} use $32$ because of memory limits.

\paragraph{Sequences, time, and splits.}
On \textbf{Gender} and \textbf{Alphabattle}, sequences are truncated or padded to at most $512$ events per forward pass; on \textbf{Age} the cap is $1024$.
Timestamps are represented as successive inter-event differences during training and mapped back to absolute time before metric computation.
After removing the held-out test users (Appendix~\ref{app:preprocessing}), we randomly assign $15\%$ of the remaining users to validation for model selection and early stopping.

\paragraph{Compute and inference calibration.}
Each model is trained on a single NVIDIA A100 80GB GPU. Hyperparameter grids, detector ablations, and other horizons require additional compute beyond the final runs in Appendix~\ref{app:runtime-h64}.
Validation loss often correlates only weakly with forecast metrics, so we select hyperparameters on the validation set by a common rule: take the configuration with the highest cDFS, or the highest TMS when cDFS differs only slightly or is low for all candidates, which is the case for most models (\S\ref{sec:results}). We apply this rule first to the learning rate, then to the sampling knob at inference: temperature for \textbf{AR} and \textbf{MT}, and classifier-free guidance scale for \textbf{LDM} and \textbf{DDM}. Exceptions: for \textbf{AR} and \textbf{LDM}, the learning rate is chosen mainly by validation loss; for the guidance scale of \textbf{LDM}, we take the highest TMS that keeps cDFS close to its best value; for the VAE of \textbf{LDM}, we take the checkpoint with the lowest KL among the two with the best reconstruction.

\paragraph{Classifier-free guidance (\textbf{LDM} and \textbf{DDM}).}
Both diffusion models use classifier-free guidance~\citep{cfg}.
During training, conditioning is dropped with probability $p_{\mathrm{uncond}} = 0.1$ so the score network sees unconditional targets as well as conditional ones.
At inference we grid-search the guidance scale over $\{0, 1, 1.2, 1.3, 1.4, 1.5, 2, 3, 4, 5\}$ as described above.

\paragraph{Feature representations and losses.}
Categorical attributes use embedding vectors of dimension $8$; numerical attributes are projected linearly to dimension $8$.
Reconstruction losses combine mean absolute error on numerical features with cross-entropy on categorical features.

\paragraph{Shared transformer hyperparameters.}
\textbf{AR}, \textbf{MT}, \textbf{LDM}, and \textbf{DDM} use the same transformer trunk: $6$ layers, $8$ attention heads, hidden size $512$, dropout probability $0.1$, and feed-forward expansion factor $4$.

\paragraph{GPT-style baselines.}
\textbf{AR} follows a GPT-2-style causal decoder; our code is adapted from the \texttt{nanoGPT} reference implementation.\footnote{\url{https://github.com/karpathy/nanoGPT/blob/master/model.py}}
Unlike the other sequence models, \textbf{AR} is trained without sampling a random history endpoint; random cutoffs are used only at evaluation.
\textbf{MT} reuses the same trunk but is a \emph{non-autoregressive} multi-step head: it pools the final hidden state and applies a linear projection to obtain $L$ parallel next-event predictions in one pass (no autoregressive roll-out over future time steps).

\paragraph{$\beta$-VAE encoder (TabSyn-style).}
Events are encoded independently with the variational architecture from TabSyn~\citep{tabsyn}: $3$ layers, $1$ attention head per feature, feature embedding dimension $8$, and feed-forward expansion factor $32$.
The KL weight follows a scheduled $\beta$: initial value $0.01$, multiplicative decay factor $0.7$, patience five validation checks, and floor $10^{-5}$.

\paragraph{DEF.}
DEF~\citep{def}; we adapt the publicly released implementation.\footnote{\url{https://github.com/ivan-chai/hotpp-benchmark}}
The history encoder is a single-layer GRU with hidden width $512$, following the reference training setup.
We use a loss subset ratio of $0.05$, matching weights $\{\texttt{\_presence} : 4\}$, and $k_{\mathrm{factor}} = 3$ as recommended in the original work.
Where the baseline method applies a presence head to keep events inside a prescribed forecast horizon, we instead use its scores to retain the top-$L$ candidates.
Our implementation predicts the full heterogeneous feature vector at each retained event, rather than only the time and a single mark.

\paragraph{DDM.}
We run the reverse process for $500$ diffusion steps.
The history is encoded with a one-layer GRU (hidden width $512$).
We follow the original implementation of \citet{cdiff} except that we disable the multiple random time draws used in the original training recipe and keep a single sampling schedule.
The original work also aggregates five forecasts at inference, averaging inter-arrival times and taking a majority vote over marks. We drop this step so that \textbf{DDM} measures the generative model alone, and treat sample aggregation as a separate axis evaluated through the Wasserstein BaryBooster (\S\ref{sec:barybooster}); \textbf{DDM-WB} is the aggregated variant. Averaging and majority voting also do not extend to the numerical features of general event sequences, whereas the barycentric aggregation does.

\paragraph{LDM.}
We use $50$ diffusion steps in the latent space.
The denoiser follows a Diffusion Transformer (DiT) layout as in ADiff4TPP~\citep{adiff4tpp}, trained with the elucidated diffusion objective of \citet{karras2022elucidating}.
A one-layer GRU (hidden width $512$) produces the history embedding consumed by the transformer.
Inside the denoiser, we further condition on the $\beta$-VAE latents of the last $L$ history events by prepending them to the noised latent sequence being denoised.

\section{Runtime and memory consumption}
\label{app:runtime-h64}

This appendix summarizes training time, inference time, and post-evaluation GPU memory usage for the main learned models at horizon $L=64$. Memory is reported in GB.
Inference times were measured under a single controlled protocol: the full test split, one job per GPU. Training times were logged on a shared cluster where a GPU could host several concurrent experiments; they should be read as upper bounds, with the relative comparison remaining fair since all models were trained under the same scheduling conditions.

\begin{table}[ht]
\centering
\begin{tabular}{llrrr}
\toprule
Dataset & Model & Train (min) & Inference (min) & Memory (GB) \\
\midrule
Age & LDM & 525.6 & 1.2 & 1.71 \\
Age & DDM & 776.9 & 8.0 & 1.62 \\
Age & DEF & 281.0 & 0.13 & 71.12 \\
Age & MT & 111.9 & 0.08 & 3.94 \\
Age & AR & 83.1 & 2.4 & 7.43 \\
Age & LDM-WB ($N{=}30$) & -- & 52.4 & 1.71 \\
\midrule
Gender & LDM & 219.7 & 0.9 & 1.08 \\
Gender & DDM & 94.6 & 7.0 & 1.22 \\
Gender & DEF & 84.3 & 0.08 & 36.84 \\
Gender & MT & 14.4 & 0.05 & 1.93 \\
Gender & AR & 14.7 & 1.0 & 3.94 \\
Gender & LDM-WB ($N{=}30$) & -- & 44.0 & 1.08 \\
\midrule
Alphabattle & LDM & 1844.3 & 5.4 & 3.35 \\
Alphabattle & DDM & 807.4 & 34.0 & 1.25 \\
Alphabattle & DEF & 473.0 & 0.4 & 36.68 \\
Alphabattle & MT & 116.3 & 0.2 & 1.94 \\
Alphabattle & AR & 83.6 & 5.5 & 3.95 \\
Alphabattle & LDM-WB ($N{=}30$) & -- & 243.7 & 3.35 \\
\bottomrule
\end{tabular}
\caption{Runtime and memory summary for Age, Gender, and Alphabattle at horizon $L=64$.}
\label{tab:runtime-h64}
\end{table}

Table~\ref{tab:booster-profile} decomposes the BaryBooster wall time
($N=30$, 10 test batches per dataset) into the base-model sampling and the
stages of the algorithm itself (medoid selection, Hungarian matching,
aggregation). Sampling dominates at both horizons (${\approx}85\%$ at $L=32$,
${\approx}62\%$ at $L=64$); the share of the matching grows with the horizon,
since the Hungarian step scales faster with the continuation length than
sampling does. Note that the $N$ continuations are sampled independently, so the dominant
stage can be distributed across several GPUs, reducing the time of this stage almost
proportionally.

\begin{table}[ht]
\centering
\begin{tabular}{llrrrrr}
\toprule
Dataset & Horizon & Total (s) & Sampling & Medoid & Matching & Aggregation \\
\midrule
Age         & 32 & 310.0 & 264.9 (85\%) & 17.8 & 26.2 & 0.3 \\
Gender      & 32 & 307.2 & 261.8 (85\%) & 18.0 & 26.4 & 0.3 \\
Alphabattle & 32 & 349.9 & 301.6 (86\%) & 18.2 & 29.0 & 0.3 \\
\midrule
Age         & 64 & 788.7 & 484.8 (61\%) & 108.0 & 194.8 & 0.2 \\
Gender      & 64 & 786.3 & 480.6 (61\%) & 107.5 & 197.3 & 0.2 \\
Alphabattle & 64 & 861.5 & 547.2 (64\%) & 105.6 & 207.7 & 0.3 \\
\bottomrule
\end{tabular}
\caption{Decomposition of the BaryBooster wall time into base-model sampling and the stages of the aggregation algorithm. Seconds per 10 test batches, $N=30$.}
\label{tab:booster-profile}
\end{table}

%%%%%%%%%%%%%%%%%%%%%%%%%%%%%%%%%%%%%%%%%%%%%%%%%%%%%%%%%%%%

\section{Public datasets considered}\label{app:datasets-rejected}

We surveyed datasets used by the closest prior benchmarks and adjacent generative-modeling papers for event sequences --- EasyTPP~\citep{easytpp}, HoTPP~\citep{hotpp}, CDiff~\citep{cdiff}, ADiff4TPP~\citep{adiff4tpp}, TabDiT~\citep{tabdit}, EBES~\citep{ebes}, and \citet{surkov2025humanactions} --- and considered each public corpus against two quantitative criteria induced by our protocol and five qualitative ones induced by the task definition.

\paragraph{Quantitative thresholds.}
We drop any candidate with strictly fewer than $5{,}000$ sequences as \emph{too small} --- a fixed cutoff that is independent of our benchmark construction (Table~\ref{tab:datasets}).
Our forecast horizons go up to $L=32$ in the main results and $L=64$ in the long-horizon study, so a usable corpus must contain at least $\text{history}+L \ge 64$ events on average; we drop any candidate with \emph{mean sequence length} below $64$ as \emph{too short}.

\paragraph{Qualitative criteria.}
\textbf{(i) Repetitive identical events.} On corpora dominated by long runs of literally identical events, the trivial \textbf{Repeat} baseline scores high on TMS by construction and a discriminator cannot meaningfully separate continuations, collapsing both axes of our evaluation. This applies to \emph{Taobao} (in the EBES and \citet{surkov2025humanactions} preprocessing), \emph{Retail / X5 RetailHero}, and \emph{Megamarket}.
\textbf{(ii) Recommendation-scale cardinality.} Datasets such as \emph{Megamarket} and \emph{Zvuk} have item or artist vocabularies in the $10^{4}\text{--}10^{5}$ range and live in the recommender-systems literature, where the standard scores (HitRate@$K$, NDCG@$K$, MRR) and modeling assumptions differ substantially from the generative-forecasting setting we target.
\textbf{(iii) Order semantics broken at source.} The Multimodal Banking Dataset (\emph{MBD}) is released with shuffled event order and is therefore incompatible with any temporal-order-aware metric or model.
\textbf{(iv) Continuous time series mislabeled as event sequences.} \emph{PhysioNet 2012} is essentially an irregularly-sampled multivariate clinical time series with mostly missing values; it is a missing-data interpolation problem, not a heterogeneous event-sequence forecasting one.
\textbf{(v) Large number of events with identical timestamps.} In several corpora, many events share the same timestamp within a sequence, weakening temporal ordering signal for point-process style forecasting. This is prominent in \emph{MIMIC-IV}~\citep{mimic4}, \emph{Retail / X5 RetailHero}, \emph{Taobao} (User Behavior), \emph{Zvuk}, and \emph{Megamarket}.

\paragraph{Catalog.}
Table~\ref{tab:datasets-rejected} consolidates the corpora we encountered. For datasets that appear in multiple papers under different names but share a single underlying source --- HoTPP \emph{Transactions}, EBES \emph{AGE}, TabDiT \emph{Age1}, and \citet{surkov2025humanactions} \emph{AGE} are the same Sberbank Sirius bank-transaction corpus; \emph{Alphabattle-2.0} in \citet{surkov2025humanactions} is the same source as our \textbf{Alphabattle}, with us using the official test split of the Alfa-Bank release --- we report a single canonical row. Counts and mean lengths are taken from each paper's own statistics tables (HoTPP Table~1; EBES Table~1; TabDiT Appendix~E Table~9; \citet{surkov2025humanactions} Table~1) and from the EasyTPP HuggingFace mirrors when paper-side numbers are ambiguous (e.g., when EasyTPP reports event-token counts rather than sequence counts).

After applying these filters, the only public corpora that satisfy both quantitative thresholds ($\geq\!5{,}000$ sequences and mean length $\geq 64$) are bank-transaction event sequences with heterogeneous fields, and have unshuffled temporal order are the Sberbank Sirius corpus (= our \textbf{Age}) and the Alfa-Bank Alphabattle release (= our \textbf{Alphabattle}). Even though lowering the size bar to $5\mathrm{k}$ leaves many more corpora above the sequence-count cutoff, nearly all of them still fail the mean-length requirement or the qualitative criteria above; our three benchmark datasets (\textbf{Gender}, \textbf{Age}, \textbf{Alphabattle}) each exceed $5\mathrm{k}$ sequences (Table~\ref{tab:datasets}).

\begin{table}[H]
\centering
\footnotesize
\setlength{\tabcolsep}{4pt}
\renewcommand{\arraystretch}{1.0}
\begin{tabular}{@{}l >{\raggedright\arraybackslash}p{3.cm} r r c@{}}
\toprule
Dataset & Reported by & \# seq. & Mean len. & Reason \\
\midrule
\multicolumn{5}{@{}l}{\emph{Our benchmark (Table~\ref{tab:datasets})}} \\
\midrule
\textbf{Gender}       & ours                                               & 14k (11k/2.8k)   & 404     & \textbf{we use} \\
\textbf{Age}          & ours                                               & 30k (24k/6k)     & 881     & \textbf{we use} \\
\textbf{Alphabattle}  & ours (Alfa-Bank, official test split)              & 135k (100k/35k)  & 367     & \textbf{we use} \\
\midrule
\multicolumn{5}{@{}l}{\emph{Generative-modeling papers for event sequences}} \\
\midrule
Retweet               & \citep{easytpp,hotpp,cdiff,adiff4tpp}              & 24k     & 41--56  & L \\
Amazon                & \citep{easytpp,hotpp,cdiff,adiff4tpp}              & 9k      & 44      & L \\
StackOverflow         & \citep{easytpp,hotpp,cdiff,adiff4tpp}              & 2k      & 64      & S, L \\
Taxi                  & \citep{easytpp,cdiff,adiff4tpp}                    & 2k      & 37--39  & S, L \\
Taobao (TPP cut)      & \citep{easytpp,cdiff,adiff4tpp}                    & 2--5k   & 58--150 & S, R \\
MOOC                  & \citep{cdiff}                                      & 4k      & n/a     & S \\
MIMIC-IV              & \citep{hotpp,mimic4}                               & 120k    & 19.7    & L, I \\
Age2                  & \citep{tabdit}                                     & 43k     & 84      & X \\
Leaving               & \citep{tabdit}                                     & 5k      & 98      & S \\
PKDD'99 Financial     & \citep{tabdit}                                     & 4.5k    & 235     & S \\
Rossmann              & \citep{tabdit}                                     & 1.1k    & 61      & S, L \\
Airbnb                & \citep{tabdit}                                     & 10k     & 19      & L \\
Large-Scale Bank Data & \citep{tabdit}                                     & 110k    & 873     & P \\
\midrule
\multicolumn{5}{@{}l}{\emph{Discriminative event-sequence benchmarks (EBES) and adjacent}} \\
\midrule
ElectricDevices       & \citep{ebes}                                       & 17k     & 96      & T \\
ArabicDigits          & \citep{ebes}                                       & 9k      & 40      & L, T \\
PhysioNet 2012        & \citep{ebes}                                       & 8k      & 75      & T \\
MIMIC-III             & \citep{ebes}                                       & 56k     & 58      & L$^\dagger$, T \\
Pendulum (synthetic)  & \citep{ebes}                                       & 100k    & 32      & L, Y \\
BPI17                 & \citep{ebes}                                       & 43k     & 13      & L \\
Retail / X5 RetailHero& \citep{ebes,surkov2025humanactions}                & 40--399k & 112--114 & R, I \\
Taobao (User Behav.)  & \citep{ebes,surkov2025humanactions}                & 10k     & 535     & R, I \\
MBD                   & \citep{ebes,surkov2025humanactions}                & 1.5--9.2M & 21--313 & O \\
Megamarket            & \citep{surkov2025humanactions}                     & 2.73M   & 653     & C, R, I \\
Zvuk                  & \citep{surkov2025humanactions}                     & 380k    & 1020    & C, I \\
Shakespeare           & \citep{surkov2025humanactions}                     & 5k      & 106     & T \\
Hawkes-1D             & \citep{easytpp,cdiff}                              & 77k     & 79      & Y \\
\midrule
\multicolumn{5}{@{}l}{\emph{Same source as our datasets (already covered above)}} \\
\midrule
Transactions / AGE / Age1 & \citep{hotpp,ebes,tabdit,surkov2025humanactions} & 24--50k & 875--882 & = \textbf{Age} \\
Alphabattle-2.0       & \citep{surkov2025humanactions}                     & 1M      & 213     & = \textbf{Alphabattle} \\
\bottomrule
\end{tabular}

\smallskip
\noindent\footnotesize
\textbf{Reason key.}
\textbf{S} = too small ($<\!5\mathrm{k}$ sequences);
\textbf{L} = too short (mean length $<\!64$);
\textbf{R} = long runs of repetitive identical events (trivializes \textbf{Repeat} and cDFS);
\textbf{I} = large number of events with identical timestamps;
\textbf{C} = recommender-systems scale cardinality (item/artist IDs $10^{4}\!-\!10^{5}$);
\textbf{O} = released with shuffled event order;
\textbf{T} = continuous / regular time series, not heterogeneous EvS (or text);
\textbf{P} = not publicly released;
\textbf{Y} = synthetic;
\textbf{X} = redundant lineage with our \textbf{Age};
\textbf{$\dagger$} = borderline (within $\sim\!10\%$ of the cutoff).
\caption{Public event-sequence datasets considered and reasons for exclusion. }
\label{tab:datasets-rejected}
\end{table}

\section{Transfer to non-transactional TPP corpora}\label{app:tpp}

To probe how the conclusions transfer beyond card transactions, we
evaluate the benchmark on two public TPP corpora, Retweet and
Amazon~\citep{easytpp}. Both fail our selection criteria
(Appendix~\ref{app:datasets-rejected}): they are much smaller, with far shorter
sequences, and each event carries only a timestamp and a categorical mark.
This makes them a stress test rather than an extension of the benchmark: if the
findings survive even there, they are not an artifact of the transaction
domain. Since these corpora carry no amount feature, TMS is computed without
the amount term. Table~\ref{tab:tpp} reports the full pool of methods at
horizon $L=32$, with Age alongside for reference.

The main findings carry over despite the harder training conditions:
the accuracy--realism trade-off holds, LDM remains the most realistic model on
both corpora, point predictors stay far below LDM on cDFS, and
LDM-WB attains the best TMS on both datasets while retaining a large part of
the realism (cDFS 0.672 on Retweet and 0.243 on Amazon vs. 0.017 and 0.002 for
DEF).

\begin{table}[ht]
\centering
\begin{tabular}{lcccccc}
\toprule
Method & \multicolumn{3}{c}{TMS $\uparrow$} & \multicolumn{3}{c}{cDFS $\uparrow$} \\
\cmidrule(lr){2-4} \cmidrule(lr){5-7}
 & Age & Retweet & Amazon & Age & Retweet & Amazon \\
\midrule
Hist. & 0.366 & 0.391 & 0.343 & \underline{0.234} & 0.035 & 0.082 \\
Mode & 0.263 & 0.156 & 0.036 & 0.000 & 0.009 & 0.002 \\
Rep. & 0.454 & 0.435 & \dashuline{0.415} & 0.065 & 0.053 & 0.070 \\
AR & 0.354 & 0.530 & 0.151 & 0.000 & 0.041 & 0.110 \\
MT & 0.360 & \dashuline{0.604} & 0.196 & 0.000 & \dashuline{0.441} & 0.140 \\
LDM & 0.426 & 0.579 & 0.357 & \textbf{1.000} & \textbf{1.000} & \textbf{0.608} \\
DDM & 0.441 & 0.565 & 0.348 & 0.202 & 0.192 & \underline{0.557} \\
DEF & \underline{0.508} & \underline{0.653} & \underline{0.430} & 0.000 & 0.017 & 0.002 \\
\midrule
LDM-WB & \textbf{0.522} & \textbf{0.663} & \textbf{0.438} & \dashuline{0.224} & \underline{0.672} & 0.243 \\
DDM-WB & \dashuline{0.480} & 0.566 & 0.412 & 0.102 & 0.144 & \dashuline{0.428} \\
\bottomrule
\end{tabular}
\caption{TMS and cDFS at horizon $L=32$ on the two TPP corpora, with Age for reference. Within each column: \textbf{best}, \uline{second}, \dashuline{third}.}
\label{tab:tpp}
\end{table}

\newpage
\section*{NeurIPS Paper Checklist}

\begin{enumerate}

\item {\bf Claims}
    \item[] Question: Do the main claims made in the abstract and introduction accurately reflect the paper's contributions and scope?
    \item[] Answer: \answerYes{}
    \item[] Justification: The abstract and Introduction state the same scope, the benchmark setup, the empirical trade-off, and Wasserstein BaryBooster; the contribution list points to Sections~\ref{sec:methods:baselines}--\ref{sec:def}, \ref{sec:benchmark}, \ref{sec:results}, and \ref{sec:barybooster} where these are developed.

\item {\bf Limitations}
    \item[] Question: Does the paper discuss the limitations of the work performed by the authors?
    \item[] Answer: \answerYes{} % Replace by \answerYes{}, \answerNo{}, or \answerNA{}.
    \item[] Justification: See \S\ref{sec:discussion:limitations}

\item {\bf Theory assumptions and proofs}
    \item[] Question: For each theoretical result, does the paper provide the full set of assumptions and a complete (and correct) proof?
    \item[] Answer: \answerNA{}
    \item[] Justification: The paper is empirical; it does not present numbered theoretical results with explicit assumption blocks and complete proofs.

    \item {\bf Experimental result reproducibility}
    \item[] Question: Does the paper fully disclose all the information needed to reproduce the main experimental results of the paper to the extent that it affects the main claims and/or conclusions of the paper (regardless of whether the code and data are provided or not)?
    \item[] Answer: \answerYes{}
    \item[] Justification: Methods, metrics (including BaryBooster, Appendix~\ref{app:barybooster}), datasets, splits, and training/evaluation protocol are specified in the main text and Appendices~\ref{app:preprocessing}--\ref{app:methods}; public data URLs appear in \S\ref{sec:datasets}.

\item {\bf Open access to data and code}
    \item[] Question: Does the paper provide open access to the data and code, with sufficient instructions to faithfully reproduce the main experimental results, as described in supplemental material?
    \item[] Answer: \answerYes{}
    \item[] Justification: The full source code, with documentation, is publicly available (\S\ref{sec:benchmark}) and enables reproduction of the experiments.

\item {\bf Experimental setting/details}
    \item[] Question: Does the paper specify all the training and test details (e.g., data splits, hyperparameters, how they were chosen, type of optimizer) necessary to understand the results?
    \item[] Answer: \answerYes{}
    \item[] Justification: Core choices appear in \S\ref{sec:results} and \S\ref{sec:benchmark}; full optimization, architectures, splits, and schedules are in Appendix~\ref{app:methods} and \ref{app:preprocessing}.

\item {\bf Experiment statistical significance}
    \item[] Question: Does the paper report error bars suitably and correctly defined or other appropriate information about the statistical significance of the experiments?
    \item[] Answer: \answerYes{}
    \item[] Justification: We address statistical significance in Appendix~\ref{app:more-res:seeds}.

\item {\bf Experiments compute resources}
    \item[] Question: For each experiment, does the paper provide sufficient information on the computer resources (type of compute workers, memory, time of execution) needed to reproduce the experiments?
    \item[] Answer: \answerYes{}
    \item[] Justification: Appendix~\ref{app:shared} states single NVIDIA A100 training; Appendix~\ref{app:runtime-h64} tabulates train/inference wall times and GPU memory per model/dataset while noting further compute for grids and screening (Appendix~\ref{app:shared}).
    
\item {\bf Code of ethics}
    \item[] Question: Does the research conducted in the paper conform, in every respect, with the NeurIPS Code of Ethics \url{https://neurips.cc/public/EthicsGuidelines}?
    \item[] Answer: \answerYes{}
    \item[] Justification: Secondary analysis on public anonymized benchmark corpora cited in \S\ref{sec:datasets}; no human-subjects experimentation.

\item {\bf Broader impacts}
    \item[] Question: Does the paper discuss both potential positive societal impacts and negative societal impacts of the work performed?
    \item[] Answer: \answerYes{}
    \item[] Justification: See \S\ref{sec:discussion:impacts}.
    
\item {\bf Safeguards}
    \item[] Question: Does the paper describe safeguards that have been put in place for responsible release of data or models that have a high risk for misuse (e.g., pre-trained language models, image generators, or scraped datasets)?
    \item[] Answer: \answerNA{}
    \item[] Justification: We study evaluation on curated public transactional benchmarks rather than scraped media; the released code exposes methodology but not frontier-scale multimodal generators, so no gated-release safeguards apply.

\item {\bf Licenses for existing assets}
    \item[] Question: Are the creators or original owners of assets (e.g., code, data, models), used in the paper, properly credited and are the license and terms of use explicitly mentioned and properly respected?
    \item[] Answer: \answerYes{}
    \item[] Justification: Yes. All datasets used have accompanying URLs and are explicitly cited; all codebases and models employed are properly licensed, and their licenses are respected and listed in the repository README.

\item {\bf New assets}
    \item[] Question: Are new assets introduced in the paper well documented and is the documentation provided alongside the assets?
    \item[] Answer: \answerYes{}
    \item[] Justification: Benchmark metrics and BaryBooster are defined in the main text and Appendices~\ref{app:tms}--\ref{app:barybooster}; consolidated training documentation appears in Appendix~\ref{app:methods} with the public code repository referenced in \S\ref{sec:benchmark}.

\item {\bf Crowdsourcing and research with human subjects}
    \item[] Question: For crowdsourcing experiments and research with human subjects, does the paper include the full text of instructions given to participants and screenshots, if applicable, as well as details about compensation (if any)? 
    \item[] Answer: \answerNA{}
    \item[] Justification: No crowdsourcing or primary human-subjects studies were conducted.

\item {\bf Institutional review board (IRB) approvals or equivalent for research with human subjects}
    \item[] Question: Does the paper describe potential risks incurred by study participants, whether such risks were disclosed to the subjects, and whether Institutional Review Board (IRB) approvals (or an equivalent approval/review based on the requirements of your country or institution) were obtained?
    \item[] Answer: \answerNA{}
    \item[] Justification: No IRB-governed human-subjects component; only secondary use of public datasets.

\item {\bf Declaration of LLM usage}
    \item[] Question: Does the paper describe the usage of LLMs if it is an important, original, or non-standard component of the core methods in this research? Note that if the LLM is used only for writing, editing, or formatting purposes and does \emph{not} impact the core methodology, scientific rigor, or originality of the research, declaration is not required.
    %this research? 
    \item[] Answer: \answerNA{}
    \item[] Justification: Large language models are not part of the proposed forecasting methods; any writing assistance is incidental to the technical contribution.

\end{enumerate}

\end{document}